\documentclass[10pt,journal,compsoc]{IEEEtran}
\usepackage[nocompress]{cite}
\usepackage[pdftex]{graphicx}
\usepackage{lineno}
\usepackage{caption}
\usepackage{makecell}
\usepackage{upgreek}
\usepackage{multirow}
\usepackage{mathtools}
\usepackage{mathrsfs}
\usepackage{float}
\usepackage{placeins}
\usepackage{threeparttable}
\usepackage{booktabs} % To thicken table lines
\usepackage{amssymb} 
\usepackage{pifont} 
\usepackage{enumitem}
\usepackage[T1]{fontenc}
\usepackage[normalem]{ulem}
\usepackage{ragged2e}
\usepackage[colorlinks=true,linkcolor=black,anchorcolor=black,citecolor=black,filecolor=black,menucolor=black,runcolor=black,urlcolor=black]{hyperref}
\usepackage[hyphenbreaks]{breakurl}

\DeclareMathOperator*{\argmin}{argmin}

\begin{document}

\title{Learning to See Through Dazzle}
\author{Xiaopeng Peng, Erin F. Fleet, Abbie T. Watnik, Grover A. Swartzlander, Jr.
\IEEEcompsocitemizethanks{\IEEEcompsocthanksitem Xiaopeng Peng and Grover A. Swartzlander are with the Department
of Imaging Science, Rochester Institute of Technology, Rochester,
NY, 14623.\protect\\
E-mail: {xiaopeng.peng, gaspci}@rit.edu
\IEEEcompsocthanksitem Erin F. Fleet and Abbie T. Watnik are with U.S. Naval Research Laboratory.Washington DC, 20375\protect\\
E-mail: {erin.fleet, abbie.watnik}@nrl.navy.mil}% <-this % stops a space
\thanks{}}

\IEEEtitleabstractindextext{%

\begin{abstract}
 \justifying{Machine vision is susceptible to laser dazzle, where intense laser light can blind and distort its perception of the environment through oversaturation or permanent damage to sensor pixels. Here we employ a wavefront-coded phase mask to diffuse the energy of laser light and introduce a sandwich generative adversarial network (SGAN) to restore images from complex image degradations, such as varying laser-induced image saturation, mask-induced image blurring, unknown lighting conditions, and various noise corruptions. The SGAN architecture combines discriminative and generative methods by wrapping two GANs around a learnable image deconvolution module. In addition, we make use of Fourier feature representations to reduce the spectral bias of neural networks and improve its learning of high-frequency image details. End-to-end training includes the realistic physics-based synthesis of a large set of training data from publicly available images. We trained the SGAN to suppress the peak laser irradiance as high as $10^6$ times the sensor saturation threshold - the point at which camera sensors may experience damage without the mask. The trained model was evaluated on both a synthetic data set and data collected from the laboratory. The proposed image restoration model quantitatively and qualitatively outperforms state-of-the-art methods for a wide range of scene contents, laser powers, incident laser angles, ambient illumination strengths, and noise characteristics.}
\end{abstract}

}

% make the title area
\maketitle
\IEEEdisplaynontitleabstractindextext
\IEEEpeerreviewmaketitle

\ifCLASSOPTIONcompsoc
\IEEEraisesectionheading{\section{Introduction}\label{sec:introduction}}
\else
\section{Introduction}
\label{sec:introduction}
\fi
Continuous advancements of laser technology have enabled the ready availability of low-cost, compact, and powerful lasers which, if misdirected toward an image sensor, may cause objectionable dazzle (e.g., sensor saturation and lens flare) or irreversible anomalies. For example, lasers can disrupt vision and mislead the tracking system of unmanned aerial vehicles \cite{steinvall2021potentialp2,steinvall2023laser,lewis2023disruptive}. Additionally, adversarial laser attacks against the sensor of autonomous or robotic vehicles have been demonstrated to significantly compromise their safety and reliability \cite{duan2021adversarial,kim2022engineering, sun2023embodied}. Lasers also present risks to sensors in mixed reality devices (e.g., video see-through head-mounted displays). These devices may advance the development of eye protection goggles \cite{quercioli2017beyond, owczarek2021virtual, deniel2022occupational, li2023mixed}, which are crucial in scientific experiments and industrial processes, such as aviation \cite{FAA2023laserincident}, manufacturing \cite{OSHA2023laserhazard}, and medical treatment \cite{malayanur2022laser}. Furthermore, laser-induced damage to consumer camera sensors has been reported during entertainment events, such as laser shows \cite{ILDA2022lasershow}.

The laser-induced saturation and damage of an imaging sensor depend on both the sensor and the laser characteristics. The damage thresholds of silicon-based imaging sensors are typically six to nine orders of magnitude higher than their saturation thresholds \cite{ruane2015reducing, ritt2019laser,theberge2022damage}. To mitigate laser-related sensor risks and image degradation, optical techniques such as wavelength multiplexing \cite{ritt2019preventing, ritt2020use}, coronagraphs \cite{ruane2013optical, ruane2014vortex, watnik2023separation}, polymer coatings \cite{swartzlander1993characteristics, dini2016nonlinear,  caillieaudeaux2024thermoset}, liquid crystals \cite{wang2016self,zhang2023advanced}, metamaterials \cite{howes2020optical, bonod2023linear}, integration time reduction \cite{lewis2019mitigation}, and smoke obscurants \cite{schleijpen2021smoke} have been investigated. 
No one approach has been found to simultaneously satisfy the desired bandwidth, response time, dynamic range, stability, and image quality characteristics.

\begin{figure*}
\includegraphics[width=\textwidth]{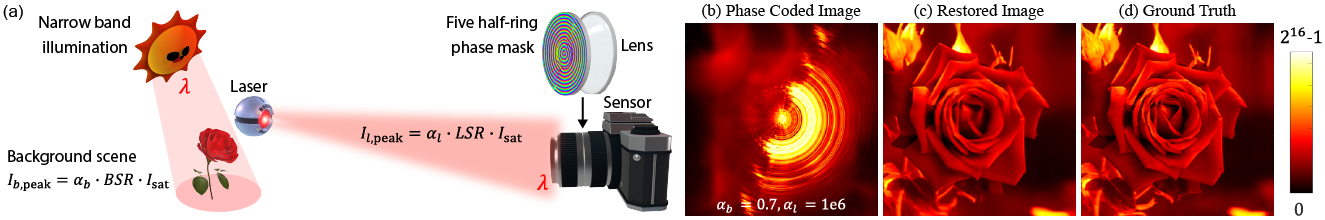}
  \caption{Laser dazzle protection at a glance. (a) Schematic of the wavefront-coded camera to diffuse laser irradiation.  Quasi-monochromatic background illumination $\lambda_b$, which is close to the laser wavelength $\lambda_b$ is assumed.  A five half-ring phase mask \cite{wirth2020half} is placed adjacent to the lens. (b) Simulation of a phase-coded image of a background scene and potential sensor-damaging laser. (c) Restored image using our SGAN-F model and (d) Ground truth image.\\}
  \label{fig:teaser}
\end{figure*}

Our approach, illustrated in
Figure \ref{fig:teaser}
is an example of computational imaging whereby a diffractive phase mask is introduced into the optical path (Figure \ref{fig:teaser}(a)) so that the ground truth scene is encoded in the recorded image in such a way to facilitate a restored image. The first task protects the sensor from laser damage and the second ensures satisfactory image quality. The phase mask is engineered to diffuse the laser light at the pupil plane, thereby reducing the peak irradiance of the focal spot while simultaneously providing adequate contrast across the transmitted spatial frequencies of the system \cite{watnik2016incoherent,ruane2016optimal,wirth2017experimental,wirth2018optimized,wirth2018computational,wirth2019point,wirth2020half, novak2020compact, novak2021imaging, peng2021cnn, peng2022computational, ghosh2023shivanet}. 
The laser, represented by a uniform planar wave front of wavelength $\lambda$, is assumed to overfill the aperture of the imaging lens at an arbitrary angle of incidence.  Here we consider that the background scene is incoherently illuminated with quasi-monochromatic light centered at the same wavelength. The phase mask, placed adjacent to the imaging lens, transforms the uncoded point spread function (PSF) into a phase-coded PSF.
Whereas the uncoded PSF produces high irradiance across a few pixels, the phase-coded PSF spreads the laser power over many hundreds of pixels, resulting in a significantly lower peak irradiance value.  The insertion of the coded phase mask blurs the scene and introduces a laser diffraction pattern on the recorded image, depicted in Figure \ref{fig:teaser}(b).

%The first task protects the sensor from laser damage and the second ensures satisfactory image quality. 

While the coded phase mask protects the sensor from being damaged by laser radiation, we face several challenges in achieving high-quality image restoration. Compared to an unprotected system, coded PSF may lead to significant image blur and saturation, as well as loss of image boundary information in recorded images. Although the PSF may be determined via simulation or measurement, the use of a single image restoration algorithm may not be sufficient to produce desirable results due to complex degradations as well as unknowns such as illumination conditions and the variable nature of the laser strength and direction. With this work, we introduce a Sandwich Generative Adversarial Network (SGAN), which wraps a learnable non-blind deconvolution module between two GANs, for the image restoration. The neural network model inpaints the laser-induced saturation, outpaints the lost image boundary areas, and reduces sensor noise through a self-attention conditional GAN, thereby producing a pre-restored image. To remove image blur caused by the use of phase mask, a learnable deconvolution is applied to a set of extracted features of the pre-restored result. The deblurred features are combined and refined with a multiscale self-attention GAN to generate the final restored image. Three variants of the SGAN model are investigated. They include the basic model SGAN-B, the enhanced model SGAN-E, and the frequency model SGAN-F. These three models make use of different feature representations and loss functions in their respective neural networks. A high-fidelity image restored by our SGAN-F is presented in Figure \ref{fig:teaser}(c), which resembles the ground truth image shown in Figure \ref{fig:teaser}(d). The specific contributions of this work are summarized as follows.

\begin{itemize}
\item A sandwich neural network architecture is introduced to restore images from laser-dazzled images, where a phase mask is employed to protect the camera from being damaged.  The model combines discriminative and generative methods to solve varying image degradations. End-to-end training encourages global optimization of the model, while the utilization of Fourier features and high-performing losses in the neural network reduces its spectral bias and further improves the image restoration accuracies. The state-of-the-art performance is validated on both numerically synthesized and experimentally acquired data.

\item Physics-based image synthesis is formulated, characterizing the real-life imaging pipeline. A large set of training images is generated numerically. Compared to experimental acquisitions, numerical synthesis obviates both hardware calibrations (e.g., image registration) and data augmentation (e.g., laser parameters, ambient illumination, and noise). Ground truth images that formed at intermediate stages and otherwise may not be able to measure in practice (e.g., the uncropped irradiance maps) are also numerically generated to supervise the neural network training.
\end{itemize}

\section{Related Work}
\label{sec:related-work}
\emph{Coded Aperture}. Computational imaging and photography are emerging areas that focus on improving and extending the capabilities of traditional imaging and camera systems using optical and computational methods \cite{barbastathis2019use,bhandari2022computational}. By altering light transmission at the pupil plane using an amplitude mask or a phase mask, the coded aperture approach has been investigated in many applications, such as coded exposure \cite{raskar2006coded}, achromatic imaging\cite{peng2016diffractive, dun2020learned},  high dynamic range \cite{sun2020learning, metzler2020deep}, lens glare suppression \cite{raskar2008glare, rouf2011glare}, extended depth of field \cite{sitzmann2018end, wu2019phasecam3d, tan2021codedstereo}, light field imaging \cite{veeraraghavan2007dappled}, lensless imaging \cite{boominathan2022recent}, privacy-preserving imaging \cite{hinojosa2021learning, tasneem2022learning}, hyperspectral imaging \cite{jeon2019compact,baek2021single}, holography \cite{peng2020neural, choi2021neural}, compressive sensing \cite{lin2014spatial, vargas2021time}, super-resolved imaging \cite{wang2018megapixel, sun2020end}, and interferemtric imaging \cite{kotwal2020interferometric, peng2017randomized, basinger2015niac}. An amplitude mask modulates the incident light by partially blocking it with a binary pattern, and the PSF is the shadow of the mask. The wavefront-coded aperture bends the incident light using a transparent phase mask, and the PSF is governed by its height profile. Compared to amplitude masks, phase elements allow higher light throughput and signal-to-noise ratio, and finer light modulation.   

Coded aperture systems rely on image reconstruction algorithms to decode high-quality images from optically coded acquisitions. Aperture masks can be designed separately or jointly optimized with the image restoration algorithms. In our approach, the PSF engineered phase mask is combined with deep learning algorithms to decode high-fidelity images from laser dazzle. The restoration model takes advantage of both discriminative and generative methods by sandwiching a learnable nonblind deconvolution between two GANs.  

\emph{Discriminative Methods}. Given a sensor image $s$ of a background scene radiance $b$, discriminative methods seek to find an estimate $\hat{b}$ through maximum-a-posteriori, which is equivalent to minimizing a regularized loss function:
\begin{equation}
\argmin_b \mathcal{L}(s - \mathcal{D}b) + \mathcal{R}(b)
\label{eq:RL}
\end{equation}

\noindent where $\mathcal{D}$ represents the degradation operator,  $\mathcal{L}(\cdot)$ and $\mathcal{R}(\cdot)$ are respectively data fidelity and regularization terms. If the degradation is assumed to be linear and the posterior has a Gaussian likelihood, Eq \ref{eq:RL} is reduced to minimizing a regularized least square loss function:
\begin{equation}
\argmin_b |s - h*b|^2 + \mathcal{R}(b)
\label{eq:RLS}
\end{equation}

\noindent where $h$ is the PSF of a shift-invariant system and the operator $*$ represents the spatial convolution. For differentiable regularization terms, the solution to Eq. \ref{eq:RLS} may be estimated by analytical inversions, such as inverse filters, Wiener deconvolution for Tikhonov regularization \cite{wiener1949extrapolation}, and fast deconvolution for hyper-Laplacian prior \cite{krishnan2009fast}. Straightforward deconvolution methods were used in early coded aperture systems \cite{raskar2006coded,talvala2007veiling,veeraraghavan2007dappled,grosse2010coded}. These approaches are sensitive to noise, and restored images may suffer from ringing artifacts. 
Alternatively, regularization terms are modeled as a known statistical distribution (e.g., heavy-tailed distribution \cite{simoncelli2001natural}, hyper-Laplacian \cite{levin2007image}, sparse gradients \cite{osher2005iterative}, etc.) or are learned through a shrinkage function \cite{schmidt2013discriminative, schmidt2014shrinkage}. Iterative solutions may be derived using splitting methods, such as half-quadratic splitting \cite{geman1995nonlinear}, reweighted least square \cite{levin2007user}, primal-dual\cite{esser2010general}, and alternating direction multiplier method \cite{boyd2011distributed}.  Iterative methods have been applied in many coded aperture systems \cite{peng2014mirror, peng2015image, monakhova2020spectral,boominathan2020phlatcam,zheng2020joint}. 

Recent advances in machine learning revitalized optimization-based restoration methods and have been applied to many coded aperture systems. Neural networks have been combined sequentially with L2-regularized deconvolution \cite{schuler2013machine, son2017fast, yanny2022deep, dong2020deep, shi2022seeing}. They have also been used in iterative deconvolution as learnable filters to encourage the sparsity of the edges of the image \cite{zhang2017learning, kruse2017learning, monakhova2019learned}. Although these methods produce sharp restored images in some cases, they may lead to erroneous results if the encouraged edges correspond to unwanted objects or artifacts. Attempts have also been made to embed DNNs into a deblur-reblur framework for self-supervised learning \cite{chen2018reblur2deblur, ren2020neural, monakhova2021untrained}. This type of approach requires accurate physics modeling of the imaging system and penalizes losses in both the forward and backward image formation pipelines. These methods are limited to stationary image formation and are susceptible to degradations of varying nature. For a small number of saturated pixels, they are discarded in the deblurring process using empirical thresholds \cite{harmeling2010multiframe,whyte2014deblurring} or a learned confidence map \cite{chen2021learning}. Sensor saturation is also approximated by differentiable functions \cite{pan2016robust,chen2021blind} and is jointly optimized with data fidelity terms.  

Learning a direct mapping from $s$ to $\hat{b}$ using DNN has been extensively studied for many low-level computer vision tasks \cite{zhang2022deep}, including image deblurring, denoising, deraining, dehazing, in- and outpainting, as well as image restoration from low light conditions. Most vision models have an encoder-decoder framework, including UNet\cite{ronneberger2015u}, residual network \cite{he2016deep}, and recurrent network \cite{tao2018scale}. The self-attention mechanism \cite{vaswani2017attention} seeks to capture long-range image dependencies.  It serves as the building blocks of vision transformers \cite{liang2021swinir,wang2022uformer,zamir2022restormer}. Its quadratic computational complexities can be reduced to linear using window partitioning techniques \cite{chen2021pre,liu2021swin} or global approximation \cite{shen2021efficient}. Self-attention has also been approximated by the spectral power density through the efficient fast Fourier transform (FFT) \cite{lee2021fnet}. As an alternative, the spatial gating unit \cite{liu2021pay} was introduced with the MLP to achieve a performance comparable to that of transformers, where similar block partition and axis swap techniques were used to reduce its complexity \cite{tu2022maxim}. Additionally, changing the image basis through linear or nonlinear transforms to
better suit the orthogonally preconditioned optimizers (e.g., ADAM, ADAgrad, etc.) was found to accelerate the convergence and improve the performance of the DNNs. One notable finding involves the use of FFT features to reduce the spectral bias of neural networks and improve its learning of high-frequency information \cite{rippel2015spectral,rahaman2019spectral, tancik2020fourier}.
Another important achievement is the use of feature matching (or perceptual) loss \cite{johnson2016perceptual} in the image intensity space, where features are extracted by pretrained VGG Net \cite{simonyan2014very} or discriminators \cite{wang2018high}. These feature representations and losses have been shown to increase the accuracy of image restoration \cite{cho2021rethinking,tsai2022stripformer, tu2022maxim, xint2023freqsel}. Furthermore, by breaking down a complicated task into subproblems and solving them progressively, multistage and multiscale frameworks allow supervision and feature fusion in multiple restoration stages and multiple image scales\cite{zhang2019deep, zamir2021multi} and encourage the recovery of image details. Embedding of kernel functions \cite{feng2021removing} and image coordinates \cite{liu2018intriguing, lin2019coco, lin2021infinitygan} into neural networks has also been introduced, respectively, to utilize the knowledge of system PSF and positional information of an image.  

\emph{Generative Methods}. Generative models seek to learn the joint distribution $\mathrm{Pr}(s,b)$. They make predictions by utilizing Bayes rules to compute the conditional probability $\mathrm{Pr}(b|s)$ and then choosing the estimate $\hat{b}$ that is the most likely to be the prediction of $b$. Modeling the joint distribution allows for more accurate recovery of missing data. Modern generative models include variational autoencoders (VAEs) \cite{kingma2014auto}, flows \cite{kingma2018glow}, generative adversarial networks (GANs) \cite{goodfellow2014generative}, and diffusion models \cite{sohl2015deep, rombach2022high}. VAEs and flows unroll the inference approximated respectively by variational and deterministic distributions, optimizing via an evidence lower bound for VAEs and the exact likelihood for flows. GANs do not rely on an explicit inference model. Instead, they learn  the target distribution from an input distribution by seeking a Nash equilibrium between a generator and a discriminator through a minimax game. Diffusion models learn the implicit latent structure of a dataset by modeling the way in which data points diffuse through the latent space. Generative methods face the performance trillema \cite{xiao2021tackling} of sampling quality, diversity, and speed. GAN models outperform many VAEs and flows in generating realistic looking images. While diffusion models can generate high-quality images with improved sample diversity \cite{dhariwal2021diffusion, ho2021classifier}, their potential in real-time applications is limited due to the high cost of iterative sampling. The single-step distilled diffusion models remain underperforming in sample quality compared to GANs\cite{zheng2023fast, meng2023distillation}.

%Such approximations tend to produce blurred results. VAEs and flows unroll the inference approximated respectively by variational and deterministic distributions, and the respect training are based on minimizing a lower bound and an exact likelihood.

GAN variations have been investigated for various applications, including image synthesis \cite{kang2023scaling, sauer2023stylegan}, domain translation \cite{isola2017image, zhu2017unpaired, gruber2019gated2depth} and adaptation \cite{shao2020domain, yang2023one}, image denoising \cite{chen2018image}, image deblurring \cite{kupyn2018deblurgan, zhang2020deblurring, pan2020physics}, image inpainting \cite{yu2018generative, yu2019free, zhao2020large} and outpainting \cite{lin2019coco, teterwak2019boundless, lin2021infinitygan, cheng2022inout}, as well as image restoration in coded aperture systems \cite{peng2019learned, rego2021robust}. Disentangled representations \cite{chen2016infogan, karras2019style, karras2021alias} may improve the performance of GAN in the generation of images of a particular style (e.g., human face, outdoor scene, etc.), and its generalization to datasets containing versatile contents may require pretrained class embedding \cite{sauer2022stylegan}. Self-attention was also introduced in GAN \cite{zhang2019self} to preserve global information in image generation. Training of a GAN can be stabilized using the gradient penalty \cite{gulrajani2017improved} or the spectrum normalization method\cite{miyato2018spectral}. 

\begin{figure*}[!htb] 
	\begin{center} 
	\includegraphics[width=1.0\linewidth]{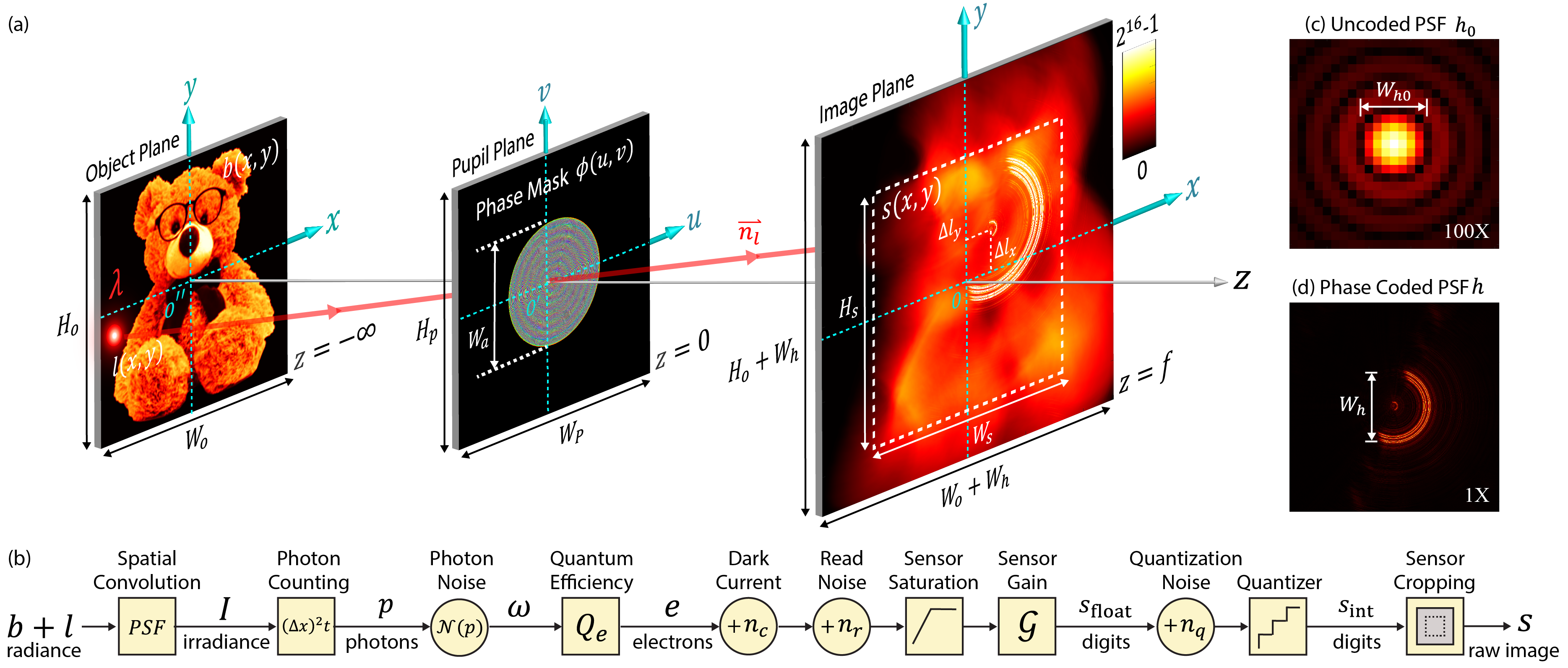} 
	\end{center} 
	\caption{Overview of the physics-based modeling of a monochromatic anti-dazzle imaging system. (a) Image formation of background scene $b$ and a laser $l$ through a phase mask and a circular lens. (b) The imaging pipeline transforms the total radiance $b+l$ into a digitized image $s$ on the sensor. The convolution of the radiance and the system PSF results in an irradiance map $I$, which determines the rate at which photons arrive at the sensor. The sensor converts photons $\omega$ to electrons $e$ given a quantum efficiency $Q_{e}$, at which stage dark current $n_c$ and read noise $n_r$ are also generated. The total number of electrons that exceeds the full well capacity may experience saturation. The electrons are then scaled by a sensor gain $\mathcal{G}$ and quantized to an array of digital counts, the dimension of which is limited by the finite sensor size ($W_s, H_s$). The simulated irradiance distribution of (c) the uncoded PSF $h_0$ and (d) the phase-coded PSF $h$.}\label{fig:image-formation}
\end{figure*}

\section{Image Formation Model}
\label{sec:imaging-formation}
Physics-based modeling allows accurate characterization of an imaging system in practice. It is an essential step towards training a data-driven restoration model with numerically simulated images and applying the trained model to the experimentally acquired data. In the following, we outline an image formation model governed by the wave propagation approach. As illustrated in Figure \ref{fig:image-formation} (a) and (b), a phase mask and a lens at the pupil plane transform the scene radiance into image irradiance. Photons are integrated over the sensor pixel during the exposure time, which are then converted to electrons and corrupted by noise. The electrons are quantized and cropped to the finite sensor size, resulting in digitized sensor images (see Figure \ref{fig:image-formation} (c)). 

\subsection{System Point Spread Function}
\label{ssec:imaging-formation-psf}
In this study, the incident laser is characterized as a plane wavefront at the entrance pupil.  For an unprotected system, the
irradiance distribution in the focal plane $(x,y)$ is described by the ideal point spread function (PSF):
\begin{equation}
h_{0}(x,y,\lambda) = \left|\frac{2J_1\left(k \rho  W_a/ f\right)}
{k \rho W_a/f} \right|^2
\label{eq:PSF_ideal}
\end{equation}
where $f$ and $W_a$ are the focal length and diameter of the imaging lens, respectively,
$k=2\pi/\lambda$ is the wave number, 
$J_1$ is the first order Bessel function of the first kind,
and $\rho = \sqrt{x^2 + y^2}$ is the radial distance
from the beam center at the Cartesian point
$(x,y) = (0,0)$.
Using the values listed in Table \ref{tab: params_phys} we obtain the diameter of the characteristic focal spot size $W_{h_0} = 2.44 \lambda f / W_a = 22$ $\upmu\mathrm{m}$ (see Figure \ref{fig:image-formation} (c)). In this study, we assume that the imaging sensor has a pixel pitch of $\Delta x = \Delta y = 5.4$ $\upmu\text{m}$, and therefore the diameter of the beam is roughly $4.1$ times the width of a single pixel.

To achieve sensor protection, we make use of a PSF that extends across many pixels (see Figure \ref{fig:image-formation} (d)),
thereby preventing the occurrence of hot spots that may dazzle or damage the sensor.  This may be achieved by introducing a coded phase element $\phi(u,v)$ 
at the entrance pupil of the system, resulting in a modified PSF:
\begin{equation}\begin{split}  
h(x,y,\lambda) = \bigg|\frac{e^{ikf}}{i\lambda f} \iint A(u,v)e^{i\phi(u,v)}e^{ik(xu+yv)/f}\mathrm{d}u\mathrm{d}v \bigg|^2
\label{eq:PSF_coded}
\end{split} 
\end{equation}

\noindent 
where $(u,v)$ is the Cartesian coordinates of pupil plane, and the optical axis coincides with the point $(u,v) = (0,0)$. To guard against numerical artifacts, we make use of a super-Gaussian aperture function $A(u,v) = \exp\big(-\big(4(u^2 + v^2)/W_a^2\big)^{50}\big)$ to represent the circular aperture. 
The numerical extent of a single pixel in the pupil plane is assigned an area $\Delta u \times \Delta v$,
and the entire numerical grid in the uv plane is assigned $N_u \times N_v$ pixels.  Thus, the physical size of the grid is $W_p \times H_p$, where $W_p = N_u \Delta u$ and $H_p = N_v \Delta v$.

The ability to successfully reconstruct an image that is blurred by the coded phase mask $\phi$ depends on the loss of contrast suffered by the modulation transfer function.  Although there is a correlation between the suppression of hot spots and the loss of contrast, one may find phase functions that lose less contrast than others, e.g, 
%Not only must the phase function spread the beam across the image plane, it must also be designed to both  allow the reconstruction of a high fidelity background scene and the elimination of obscurations caused by the laser.
the so-called ``five half-ring'' phase function \cite{wirth2020half}:\\
\begin{equation} 
\phi_{R}(\xi, \Phi) = \mathrm{atan}\bigg(\frac{1}{Q_0}
\sum_{m=1}^{17}\sum_{n=1}^{5}a_{m,n}\sin \big(m_{\mathrm{o}}(\Phi-\theta_n)\big)Q_{m,n}\bigg)
\label{eq:PhaseFunc}
\end{equation}

\noindent
where $m_o=2m+1$ and $a_{n,m}=4r_n/(m_{\mathrm{o}}\pi)$. By writing the Cartesian pupil coordinates as polar coordinates $\xi=\sqrt{u^2+v^2}$ and $\Phi=\mathrm{atan}(u/v)$, the Bessel terms may be denoted respectively as $Q_{0}(\xi)=\sum_{n=1}^5r_n J_0(2\pi\xi r_n/W_a)$ and $Q_{m,n}(\xi)=J_{2m+1}(2\pi\xi r_n/W_a)$, where $r_n$ and $\theta_n$ are the radial and azimuthal angles of each ring. A set of optimal values was reported \cite{wirth2020half}: $r_n$ = [13.6, 91.8, 6.3, 10.3, 4.2] and $\theta_n$ = [1.86, 1.09, 1.15, 1.21, 1.22] radians for $n =$ 1 to 5 respectively.  We make use of this five half-ring phase function for all the cases below.

\subsection{Sensor Image}
\label{ssec:imaging-formation-image}

\emph{Irradiance Distribution}. A shift-invariant imaging system integrates the radiance distribution over the solid angle that is extended by the aperture through spatial convolution, resulting in an irradiance map at the image plane. Here it is assumed that the background illumination has a narrow-band wavelength $\lambda_b$ and the laser has a wavelength $\lambda_l$. For a phase coded system, the irradiance distribution of the background scene and a laser are given respectively as:
\begin{subequations} \label{eq:irrad_scene} 
\begin{align} 
&I_{b}(x,y) = b(x,y) * h(x,y,\lambda_b)\\
& I_{l}(x,y) = \delta(x-\Delta l_x, y-\Delta l_y) * h(x,y,\lambda_l)
\end{align}  
\end{subequations}

\noindent where $b(x,y)$ and $l(x,y)$ are, respectively, the radiance map of the background scene and the laser. The operator $*$ denotes the spatial convolution. Here we assume the radiance maps and the sensor image are discretized by a same number of pixels, and the values of $b$ are normalized to a unit range. The laser is considered a Dirac Delta function, which targets the sensor at a normal $\vec{n}_l=(n_{u}, n_{v})$ with respect to the optical axis. Its footprint shift on the focal plane is thus given by $(\Delta l_x, \Delta l_y) = f\cdot \vec{n}_l$. By replacing the coded PSF $h$ with the uncoded PSF $h_{0}$ in Eq.\ref{eq:irrad_scene}, the irradiance maps of the background scene and the laser are defined as $I_{b0}$ and $I_{l0}$ respectively for an unprotected system.

\emph{Sensor Saturation.} For a given wavelength, the irradiance value that saturates a sensor is expressed as: 
%irradiance threshold for laser-induced sensor saturation may be expressed:
\begin{equation}
I_\mathrm{sat} (\lambda)= e_\mathrm{sat}\cdot \frac{\mathfrak{h}\cdot c}{\lambda\cdot t \cdot(\Delta x)^2 \cdot Q_{e} }
\label{eq:ThresSat}
\end{equation}
\noindent where $e_\mathrm{sat}$, $Q_{e}$, and $\Delta x$ are respectively the full well capacity, quantum efficiency, and pixel pitch of the sensor, $\mathfrak{h} = 6.63\cdot 10^{-34}$ $[\text{J} \cdot \text{s}]$ is Plank's constant,  $c = 3\cdot 10^8$ [m/s] is the speed of light in vacuum, and $t$ is the exposure time. For an unprotected system, let us express the peak irradiances of the background scene and the laser spot, respectively, as $I_{b0,\mathrm{peak}}$ and $I_{l0, \mathrm{peak}}$, which are proportionate to the irradiance saturation value:
\begin{subequations} \label{eq:peak_irrad_uncoded} 
\begin{align} 
& I_{b0,\mathrm{peak}}=\alpha_{b} \cdot I_\mathrm{sat}(\lambda_b)
    \label{eq:peak_irrad_background_uncoded} \\
& I_{l0, \mathrm{peak}}=\alpha_{l} \cdot I_\mathrm{sat}(\lambda_l)
    \label{eq:peak_irrad_laser_uncoded}
\end{align}  
\end{subequations}
where $\alpha_{b}$ and and $\alpha_l$ are respectively the strength of the background illumination and the laser. The laser saturates a single pixel when $\alpha_l = 1$ and multiple pixels when $\alpha_l > 1$.  
Sensor damage may occur at $\alpha_l > 10^6$. For an optical system protected by a pupil plane phase mask the corresponding peak irradiance values of the background scene
and laser are respectively scaled by a background suppression value $BSR$ and 
the engineered laser suppression value $LSR$:
\begin{subequations} \label{eq:peak_irrad_coded} 
\begin{align} 
& I_{b, \mathrm{peak}} = \alpha_b \cdot BSR \cdot I_\mathrm{sat}(\lambda_b)
    \label{eq:peak_irrad_background_coded} \\
& I_{l, \mathrm{peak}}=\alpha_{l} \cdot LSR \cdot I_\mathrm{sat}(\lambda_l)
    \label{eq:peak_irrad_laser_coded}
\end{align}  
\end{subequations}
\noindent
The values $BSR \sim 1$ and $LSR << 1$ lower the risk of sensor saturation and damage while maintaining the transmission rate of the scene irradiance. 
The five half-ring phase mask assumed in this report was found to have a remarkable value: $LSR = 10^{-3}$, which reduces the peak laser irradiance by three orders of magnitudes. 

\begin{table}[tb]
\centering
\captionsetup{justification=centering}
\caption{Physical Parameters}
\begin{tabular}{lcc}
\toprule
\makecell{Parameter} & \makecell{Symbol} & \makecell{Value} \\
\midrule
Background wavelength    & $\lambda_b$              & 633 nm  \\
Laser wavelength         & $\lambda_l$              & 633 nm  \\
Effective focal length   & $f$                      & 0.11 m \\
Exposure time            & $t$                      & 0.1 sec\\
Aperture diameter        & $W_a$                    & 3.83 mm \\
Quantum efficiency       & $Q_e$                    & 0.56\\
Sensor gain              & $\mathcal{G}$            & $0.37$\\
Full well capacity       & $e_\mathrm{sat}$         & $25500e^-$\\
Read noise (mean)        & $\mu_{r}$                & $390e^-$\\
Read noise (std.)        & $\sigma_{r}$             & $10.5e^-$\\
Dark current             & $\sigma_{c}$             & $0.002e^-$\\
Bit depth per channel    & $\mathrm{bpc}$           & $16$\\
Pupil pitch              & $\Delta u = \Delta v$    & 3.74 $\upmu$m\\
Sensor pitch             & $\Delta x = \Delta y$    & 5.4 $\upmu$m\\
Pupil resolution         & $N_u \times N_v$         & $4096 \times 2160$\\
Sensor resolution        & $N_x \times N_y$         & $3352 \times 2532$ \\
\bottomrule
\end{tabular}
\label{tab: params_phys}
\end{table}

\begin{figure*}[tb] 
	\begin{center} 
	\includegraphics[width=1.0\linewidth]{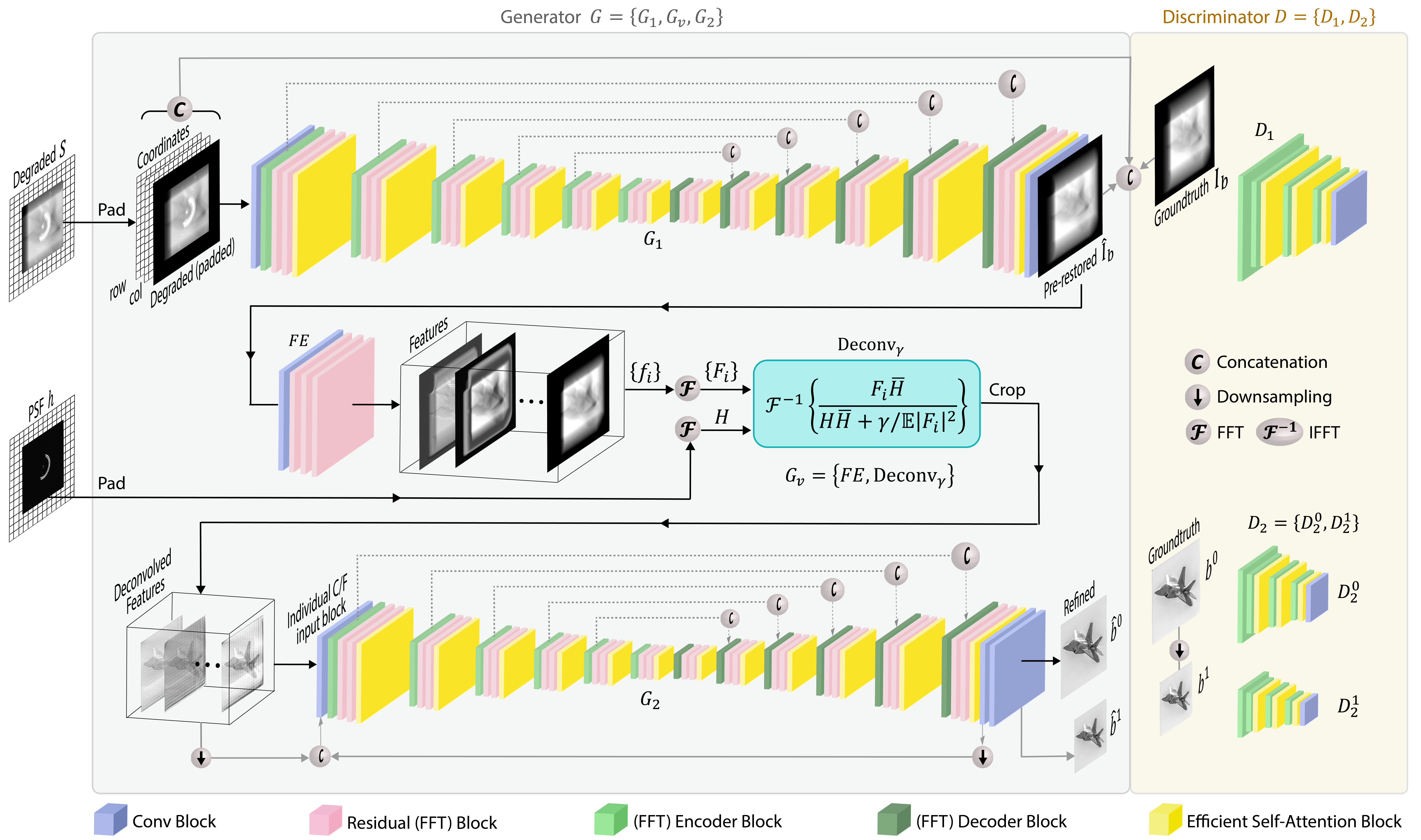} 
	\end{center} 
	\caption{The architecture of the neural sandwich GAN (SGAN) model, which consists of a set of generators  $G = \{G_1, G_\mathrm{v}, G_2\}$ and a set of discriminators $D = \{D_1, D_2\}$. The U-shape generator $G_1$ removes noise, inpaints the laser contribution, and outpaints the cutoff image boundary from the concatenation of the zero-padded input image $s$ and its coordinates $(x,y)$, producing an estimated irradiance map $\hat{I}_b$. The deconvolution module $G_v$ extracts a set of features from the pre-restored image using the feature extractor $FE$. The feature images are individually deconvolved by the Deconv engine, where the noise power spectrum $\gamma$ is learnable. The U-shape generator $G_2$ combines and refines the deconvolved features into the restored images $\{\hat{b}^L| L = 0,1\}$ on a coarse scale ($L = 0$) and a fine scale ($L = 1$) respectively. The model is adversarially trained with a conditional discriminator $D_1$ and a multiscale discriminator $D_2 = \{D_2^L\}$ end-to-end. Three variances of the SGAN are investigated. They include the basic SGAN-B model, the enhanced SGAN-E model,  and the frequency SGAN-F model. The SGAN-B/E models are built with basic residual, encode, and decoder blocks, while the SGAN-F model makes use of FFT representations in the encoders and decoders.} 
	\label{fig:Restore-Method}
\end{figure*}

\emph{Photon Counting.} Photons arrival at a sensor has a Poisson distribution, the rate of which is determined by the image irradiance $I$, the pixel pitch $\Delta x$, wavelength $\lambda$, and the integration time $t$:
\begin{equation}
p = \frac{(I_b\cdot\lambda_b + I_l\cdot\lambda_l)\cdot t\cdot(\Delta x)^2}{\mathfrak{h}\cdot c}
\label{eq:photon_count}
\end{equation}

\noindent According to the central limit theorem, the Poisson distribution may be approximated by a Gaussian distribution, which was found to be a better characterization of our sensor in practice. The Gaussian distributed photon is given by $\omega\sim \mathcal{N}(c_1\cdot \mu_{p},  c_2\cdot \sigma_{p} )$, where its mean and standard deviation are written respectively as the modulated mean ($\mu_p$) and standard deviation ($\sigma_p$) of the photon arrival rate $p$, and $c_1$ and $c_2$ are the modulation coefficients. 

\emph{Photon to Electron.} Given a quantum efficiency $Q_e$, the collected  photons are converted to electrons: $e = Q_e \cdot \omega$, followed by noise corruptions and the digitization process:
\begin{equation}
s =\mathrm{crop}\Bigg(\min\bigg(s_{\mathrm{sat}},\Big\lfloor\mathcal{G}\cdot\min\big(e_\mathrm{sat}, e + n_d + n_c\big)+ n_q\Big\rfloor\bigg)\Bigg)
\label{eq:SensorImage}
\end{equation}

\noindent where unwanted electrons generated by other factors are modeled as additive dark current $n_c$ and read noise $n_r$. The dark current has a Poisson distribution $n_c\sim \mathcal{P}(\mu_c)$, and the read noise is Gaussian distributed $n_r \sim \mathcal{N}(\mu_r, \sigma_{r})$. The mean values $\mu_c, \mu_r$ and the standard deviation $\sigma_r$ of the noise are obtained through sensor calibration (see Supplemental Information). 

\emph{Digitization.} Electrons are converted into an array of integer digital counts that represents the image recorded by the sensor. The total number of electrons that exceeds the full well capacity of the sensor $e_\mathrm{sat}$ is clipped. Electrons are then amplified by a sensor gain $\mathcal{G}$, producing an array of floating points. Uniformly distributed quantization noise $n_q\sim \mathcal{U}(-0.5, 0.5)$ is added to these digits, and floating-point digital values are then quantized to integer digital counts. The upper limit of the digital counts is determined by the bit depth per channel (bpc) of the sensor, where $\mathrm{s_{sat}} = 2^{\mathrm{bpc}}-1$.  In cases where $\mathcal{G} <e_\mathrm{sat}/\mathrm{s_{sat}}$, the quantized integer digital counts that exceed the digital upper limit are further clipped. The size of the image recorded by the sensor is determined by the finite size $(W_s, H_s)$ of that sensor. Consider that a radiance map in the object plane has a size $(W_o, H_o)$ and the system PSF has a width $W_{h}$, the size of the image formed in the focal plane is given by $(W_o+W_{h}, H_o + W_{h})$. Boundary areas that exceed the sensor size are cropped. The values of the physical parameters used in the simulation match the experiment (see Table 1).

\section{End-to-end Image Restoration}
Here we introduce a neural Sandwich GAN (SGAN) for image restoration in phase mask based anti-dazzle imaging systems. The model is designed to address the challenges of combined and dynamic degradations that are presented in these imaging scenarios. As shown in Figure \ref{fig:Restore-Method}, the architecture of the SGAN includes a learnable non-blind deconvolution module $G_v$ being wrapped between a conditional generator $G_1$ and a multi-scale generator $G_2$. Given the knowledge of the system PSF $h$, the background scene radiance $\hat{b}$ is restored from the sensor image $s$ by inverting the degradations progressively:
\begin{equation}
\begin{split}
\hat{b} = \mathrm{G}_{2}\big(G_v\big(\mathrm{G}_1(s), h\big)\big)
\end{split}
\label{eq:image_restore_model}
\end{equation}
\noindent The set of generators $\{G_1, G_v, G_2\}$ are adversarially trained with a conditional discriminator $D_1$ and a multiscale discriminator $D_2 = \{D_2^{L}| L = 0,1\}$. End-to-end training of SGAN models further encourages each solution to converge to a global optimum.

\subsection{Pre-Restoration with Conditional GAN}
\label{ssec: Method-PreRestore}

The pre-restoration GAN consists of a generator $G_1$ and a discriminator $D_1$. The generator seeks to find a mapping from the degraded sensor image $s$ to an estimated irradiance map $\hat{I}_b = G_1(s, x, y)$ conditioned on its coordinates $(x,y)$. To allow adequate support for the recovery of cut-off boundaries, the input image is zero-padded from $(W_s, H_s)$ to a minimum $(W_s + W_{h}, H_s + W_{h})$ along each dimension. The discriminator learns to distinguish whether $\hat{I}_b$ is real or fake conditioned on the degraded image and its coordinates. The objective function of Wasserstein GAN \cite{arjovsky2017wasserstein}  with gradient penalty \cite{gulrajani2017improved} is employed to stabilize the training: 
\begin{subequations}
\begin{alignat}{2}
\mathcal{L}_{GAN, G_1}(G_1, D_1) &=-\lambda_{ADV}\cdot\mathbb{E}[D_1(s,x, y, \hat{I}_b)]\\
\mathcal{L}_{GAN, D_1}(G_1, D_1) &=\lambda_{ADV}\cdot \mathbb{E}[D_1(s,x, y, I_b)] \nonumber\\
                                 &-\lambda_{ADV}\cdot\mathbb{E}[D_1(s,x, y, \hat{I}_b]\nonumber\\
                            &-\lambda_{GP}\cdot\mathbb{E}[(\lVert\nabla_{\tilde{I}_b}D_1(s,x, y, \tilde{I}_b)\rVert_2-1)^2]
\end{alignat}
\end{subequations}
\label{eq:ObjectiveFunc_CGAN}

\noindent where  $\tilde{I}_b$ is sampled uniformly along a straight line between a pair of estimated irradiance maps $\hat{I}_b$ and the ground truth irradiance maps $I_b$. In addition to minimizing the GAN objective, generators are also encouraged to produce estimates that are close to the ground truth in terms of data fidelity. The reconstruction objective is given by Charbonnier $L_1$ distance \cite{charbonnier1994two} : 
\begin{equation}
\begin{split}
\mathcal{L}_{REC1}(G_1) &= \sqrt{\lvert I_b-\hat{I}_b\rvert^2 + \epsilon}
\end{split}
\label{eq:RecLoss_CGAN}
\end{equation}

\noindent Here a residual UNet is designed for $G_1$, and the Markovian discriminator (PatchGAN) \cite{isola2017image} is employed for $D_1$ in our SGAN-B/E models. The double convolutional layers of the original UNet are replaced with residual blocks \cite{he2016deep}. The neural net blocks of this module are listed in Table \ref{tab:SGAN_G_and_D}. For the SGAN-F model, we make use of the residual FFT (RFT) blocks \cite{xint2023freqsel} in $G_1$ and FFT convolutional (FTC) blocks in $D_1$ respectively. The structure of the net blocks and the values of the hyperparameters $\lambda_{ADV}$ and $\lambda_{GP}$ are provided in the Supplementary Information.   

\subsection{Learnable Deconvolution in Feature Space}
\label{ssec: Deconv}
Given the knowledge of the system PSF $h$, the deconvolution module $G_v = \{FE, \mathrm{deconv}\}$ removes the blur introduced by the phase mask. Close-form deconvolution derived from L2 regularized least square (e.g., Wiener deconvolution \cite{wiener1949extrapolation}) provides an efficient way for non-blind deblurring. Applying deconvolution in the feature space instead of the intensity domain may improve the restoration of fine image details \cite{dong2020deep}. Using a residual net, the feature extraction ($FE$) module extracts a set of $N=16$ feature images $\{f_i = FE(\hat{I}_b)\ |\ i = 1,2, \dots, N\}$ from the latent scene irradiance map $\hat{I}_b$. Each feature image is deconvolved individually by a learnable Wiener deconvolution: 
\begin{equation}
\hat{f}_i = \mathrm{deconv}(f_i, h) = \mathcal{F}^{-1}\bigg\{\frac{F_i\bar{H}}{H\bar{H} + \gamma/\mathbb{E}|F_i|^2}\bigg\}
\label{eq:Deconvolution}
\end{equation}

\noindent where $F_i = \mathcal{F}(f_i)$ denotes the Fourier transform of each feature image, and $\mathcal{F}^{-1}$ is the inverse Fourier transform. The optical transfer function and its complex conjugate are given respectively by $H = \mathcal{F}(h)$ and $\bar{H}$. Instead of estimating a universal signal-to-noise ratio (SNR) \cite{dong2020deep, yanny2022deep} for all images, we calculate the power spectrum of each feature image $\mathbb{E}|F_i|^2$ and learn a noise power spectrum $\gamma$. It provides a more accurate estimate of the SNR for different images. In addition, the pre-restored boundary area reduces the ringing artifacts in the deconvolved feature images. 

\begin{figure*}[ht] 
	\begin{center} 
	\includegraphics[width=1.0\linewidth]{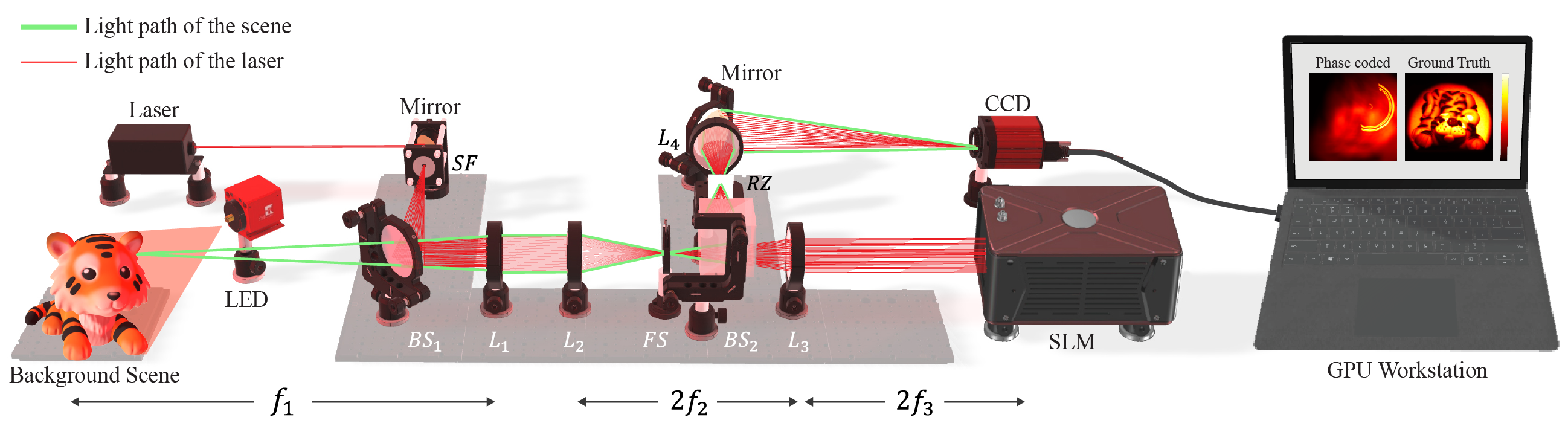} 
	\end{center} 
	\caption{Laboratory prototype of the anti-dazzle imaging system.  A coherent laser source with wavelength $\lambda=633$ nm is redirected by a mirror $M_1$ and is expanded by a spatial filter $SF$. The laser light and the incoherently illuminated background scene simultaneously pass through a beam splitter $BS_1$, forming a joint light cone. The light cone is collimated by the first lens $L_1$, which is a focal length away from the scene with $f_1 = 40$ cm. A laser line filter is attached to the light emitting diode (LED) to produce quasi-monochromatic illumination with a central wavelength $\lambda$. The pupil is located at the second lens $L_2$ with $f_2 = 10$ cm. The lens $L_3$ has a focal length $f_3 = 10$ cm and is located at $20$ cm from $L_2$. It images the pupil to the SLM, which then retro-reflects the predetermined five half-ring phase pattern to $L_3$ and produces an engineered PSF at the pupil. The coded image is formed at the focal plane of the $L_2$ where unwanted reflection is blocked by a razor blade $RZ$. The intermediate image is magnified and reimaged on to a CCD sensor by a lens $L_4$ with $f_4=20$ cm and a mirror $M_2$. A circular field stop $FS$ between $L_2$ and $BS_2$ limits the field of view. The ground truth is recorded by turning off the laser source and the SLM.}
	\label{fig:experiment-setup}
\end{figure*}

\begin{table}
	\centering
    \begin{threeparttable}
    	\caption{Generators and Discriminators in SGANs}
        \label{tab:SGAN_G_and_D}
    	\setlength{\tabcolsep}{2pt}
    	\begin{tabular}{m{50pt} m{50pt} m{50pt} m{60pt}}
    		\toprule
    		\makecell{Model}&
    		\makecell{$G_1, G_{2}$}&
    		\makecell{$FE$}&
            \makecell{$D_1, D_{2}$}\\
    		\midrule
            SGAN-B & Res UNet  & Res Net & PatchGAN\\
            SGAN-E & Res UNet & Res Net& PatchGAN\\
            SGAN-F & RFT UNet & Res Net & FTC-PatchGAN\\
            \bottomrule
    	\end{tabular}
        \begin{tablenotes}[para,flushleft]
            \footnotesize
            \vspace{1ex} Res: Residual; RFT: Residual FFT; FTC: FFT Convolution
        \end{tablenotes}
    \end{threeparttable}
\end{table}

\subsection{Refinement with Coarse-to-Fine GAN}
\label{ssec:MSGAN}
The second GAN $\{G_2, D_2\}$ seeks to find a mapping from the deconvolved features to an estimate of scene radiance. To better recover image details, a coarse-to-fine architecture is established. The deconvolved features $\{f_i\}$ are downsampled by antialiased bicubic interpolation and refined by the generator, producing estimated radiance maps $\{\hat{b}^L\} = G_2(\{f^L_i\})$, where $L = 0, 1$ represent coarse and fine scales respectively. The weights of the generator are shared across scales, except for the first two input layers which accept input features of coarse and fine scale respectively. The discriminators $D_2 = \{D^L_2 | L = 0, 1\}$ determine whether the estimate is real or false on each scale. The adversarial objective at this stage is written as: 

\begin{subequations}
\begin{alignat}{2}
\mathcal{L}_{GAN, G_2}(G_2, D_2 |G_1, G_v)&=-\lambda_{ADV}\cdot \sum_{L}\mathbb{E}[D_2^{L}(\hat{b}^L)]\\
\mathcal{L}_{GAN,D_2}(G_2, D_2 |G_1, G_v)
&= \lambda_{ADV}\cdot \sum_{L}\mathbb{E}[D_2^L(b^L)]-\mathbb{E}[D_2^{L}(\hat{b}^L)]\nonumber\\
&-\lambda_{GP}\cdot\sum_{L} \mathbb{E}[(\lVert\nabla_{\tilde{b}^L}D_2^L(\tilde{b}^L)\rVert_2-1)^2]
\end{alignat}
\end{subequations}
\label{eq:ObjectiveFunc_MSGAN}

\noindent where $\tilde{b}$ is sampled uniformly along a straight line between a pair of estimated and ground truth radiance maps $\hat{b}$ and $b$. The reconstruction objective given by the Charbonnier $L_1$ difference between the estimated and the ground truth 
radiance pyramids:
\begin{equation}
\mathcal{L}_{REC2}(G_2|G_1,G_v) = \sum_{L = 0,1}\sqrt{\lvert b^L-\hat{b}^L\rvert^2 + \epsilon}
\label{eq:RecLoss_MSGAN}
\end{equation}

\noindent where the fine-scale ground truth image is downsampled from the coarse-scale ground truth image using an anti-aliasing bicubic method \cite{hu2006adaptive}. In this method, the high-frequency components that cause aliasing artifacts are filtered by a low-pass cubic kernel. 

Similar to the pre-restoration module, FFT representations are used in the refinement module for the SGAN-F model (see Table \ref{tab:SGAN_G_and_D}). In addition, we explore the combined use of the multiscale discriminator feature matching loss (MDF)\cite{wang2018high}, VGG loss \cite{johnson2016perceptual}, and FFT loss \cite{rippel2015spectral} in three SGAN models (see Table \ref{tab:Comp_Baseline-Algo}). The MDF objective is given by:
\begin{equation}
\begin{aligned}
&\mathcal{L}_{MDF}(G_2, D_2|G_1, G_v) \\
&=\frac{1}{T_L}\sum_{L = 0,1}\sum_{t}^{T_{L}}\mathbb{E}\big[\lvert D_{2,t}^{L}(b^L)-D_{2,t}^{L}(\hat{b}^L))\rvert\big]
\end{aligned}
\label{eq:FMLoss_MSGAN}
\end{equation}

\noindent where the discriminators serve as feature extractors and do not maximize this objective during the training of $G_2$, and $T_{L}$ is the number of layers of $D_2^L$. The VGG objective is defined as:
\begin{equation}
\begin{split}
\mathcal{L}_{VGG}(G_2|G_1, G_v) = \frac{1}{T_V}\sum_{L = 0,1}\sum_{t}^{T_V}\mathbb{E}\big[\lvert V_t(b^L)-V_t\big(\hat{b}^L)\big)\rvert\big]
\end{split}
\label{eq:VGGLoss_MSGAN}
\end{equation}

\noindent where $V$ denotes the pre-trained VGG net of $T_V$ layers. The FFT objective is expressed as the sum of absolute difference between the Fourier transforms of the ground truth radiance map and estimated radiance map at fine- and coarse scales:
\begin{equation}
\begin{split}
\mathcal{L}_{FFT}(G_2|G_1, G_v) = \sum_{L = 0,1} \lvert \mathcal{F}(b^L)-\mathcal{F}\big(\hat{b}^L)\rvert
\end{split}
\label{eq:FFTLoss_MSGAN}
\end{equation}

\noindent The structure of neural network blocks and the hyperparameters $\lambda_{ADV}$ and $\lambda_{GP}$ that modulate loss functions are provided in the Supplementary Information. 

\subsection{End-to-End Training}
The SGAN models are trained end-to-end through a joint objective function, where the adversarial term is given by:
\begin{subequations}
\begin{align}
\mathcal{L}_{\mathrm{GAN, G}}(G_1, G_v, G_2, D_1, D_2) &= \mathcal{L}_{GAN, G_1}+\mathcal{L}_{GAN,G_2}\\ 
\mathcal{L}_{\mathrm{GAN, D}}(G_1, G_v, G_2, D_1, D_2) &= \mathcal{L}_{GAN, D_1}+\mathcal{L}_{GAN,D_2}
\end{align}
\end{subequations}
\label{eq:TotalObj_GAN}

\noindent The reconstruction objectives for the SGAN-B/E/F models are listed respectively as follows:
\begin{equation}
\begin{split}
\mathcal{L}_{\mathrm{REC-B}}(G_1, G_v, G_2, D_2) &= \lambda_{REC}\cdot\mathcal{L}_{REC1}(G_1) \\
&+ \lambda_{REC}\cdot\mathcal{L}_{REC2}(G_2|G_1, G_v) \\
&+ \lambda_{MDF}\cdot\mathcal{L}_{MDF}(G_2, D_2|G_1, G_v)
\end{split}
\label{eq:TotalObj_B}
\end{equation}

\begin{equation}
\begin{split}
\mathcal{L}_{\mathrm{REC-E}}(G_1, G_v, G_2, D_2) &= \mathcal{L}_{\mathrm{REC-B}}(G_1, G_v, G_2, D_2)\\ &+\lambda_{VGG}\cdot\mathcal{L}_{VGG}(G_2|G_1, G_v)\\
\end{split}
\label{eq:TotalObj_E}
\end{equation}

\begin{equation}
\begin{split}
\mathcal{L}_{\mathrm{REC-F}}(G_1, G_v, G_2, D_2) &= \mathcal{L}_{\mathrm{REC-E}}(G_1, G_v, G_2, D_2)\\
&+ \lambda_{FFT}\cdot\mathcal{L}_{FFT}(G_2|G_1, G_v)\\
\end{split}
\label{eq:TotalObj_F}
\end{equation}

\noindent The generators and the discriminators are trained in an alternating manner. Denote $L_{GAN} = L_{GAN, G}$ and $L_{GAN} = L_{GAN, D}$ as the adversarial objectives for generators and discriminators respectively, the generators $G_1, G_v, G_2$ seek to minimize the adversarial and reconstruction objectives, while the discriminators $D_1$ and $D_2$ aim to maximize only the adversarial terms:
\begin{equation}
\min_{G_1, G_v, G_2}\Bigg(\bigg(\max_{D_1, D_{2}}\mathcal{L}_{\mathrm{GAN}}\bigg) + \mathcal{L}_{\mathrm{REC}}\Bigg)
\label{eq:TotalObj_Solution}
\end{equation}

\noindent Using only the generators at the inference stage, the radiance of the scene is restored through Eq. \ref{eq:image_restore_model}. The values of the hyperparameters $\lambda_{REC}$, $\lambda_{VGG}$, $\lambda_{FFT}$ are provided in Supplemental Information. 

\subsection{Efficient Self-Attention}
\label{ssec:ATTN}
The building blocks of our SGAN model are convolutional layers. The convolution processes the information in a local neighborhood, which limits the neural net from capturing long-range dependencies in an image. To model global dependencies between widely separated spatial regions, a self-attention block is inserted in each layer of $\{G_1, G_2\}$ and $\{D_1, D_2\}$. The block consists of three linear layers $M_1, M_2, M_3$, which embed the vectorized latent feature $f\in \mathbb{R}^{d_n}$ into three distinct feature spaces: $Q = M_1(f)\in \mathbb{R}^{d_n\times d_k}$, $K = M_2(f)\in \mathbb{R}^{d_n\times d_k}$, and $V = M_3(f)\in \mathbb{R}^{d_n\times d_n}$.  The computation of self-attention $D(Q,K,V) = \mathrm{softmax} (QK^{\mathrm{T}})V$ has a quadratic complexity $\mathcal{O}(d_n^2d_k)$, which limits its use in high-resolution images. Here we adopt an efficient approximation scheme \cite{shen2021efficient}, which allows the computation of global self-attention with linear complexities. The normalizations are applied respectively to the row and column vectors of $Q$ and $K$:
\begin{equation}
E(Q,K,V) = \mathrm{softmax}(Q_\mathrm{row})\big[\mathrm{softmax}(K_\mathrm{col}^{\mathrm{T}})V\big]\
\label{eq:EffAttn}
\end{equation}

\noindent Changing the order of matrix multiplications by the commutative property reduces the complexity to linear $\mathcal{O}(d_k^2d_n)$, here $d_k = 8$. The output feature is given by $f_\mathrm{attn} = R[E(Q,K,V)\big] + f$, where $R$ is a linear layer initialized to zeros to encourage gradual learning of global evidence.

\begin{table*}
	\centering
    \begin{threeparttable}
    	\caption{Comparisons of Image Restoration Algorithms for Anti-Dazzle Imaging}
        \label{tab:Comp_Baseline-Algo}
    	\setlength{\tabcolsep}{2pt}
    	\begin{tabular}{m{75pt} m{122pt} m{188pt} m{100pt}}
    		\toprule
    		\makecell{Model}&
    		\makecell{Model Type}&
    		\makecell{Strategies}&
            \makecell{Loss Functions}\\
    		\midrule
            Pix2Pix \cite{isola2017image}& 
            UNet CGAN & 
            -& 
            L1, GAN\\
            DeblurGAN \cite{kupyn2018deblurgan}& 
            REDNet CGAN & 
            - & 
            VGG, GAN\\
            ST-CGAN \cite{wang2018stacked}& 
            Stacked UNet CGAN& 
            Multi-stage& 
            L1, GAN\\
            WienerNet \cite{yanny2022deep}& 
            L2 Deconv + UNet& 
            Le-SNR& L1, SSIM \\
            DeepWiener \cite{dong2020deep}& 
            L2 F-Deconv + REDNet& 
            multiscale& 
            L1 \\
            MPRNet \cite{zamir2021multi} & 
            Multi-stage UNet& 
            Multi-stage, multi-patch, CA, CSFF, SAM & 
            L1, Edge\\
            MIMO-UNet \cite{cho2021rethinking} & 
            MIMO UNet with Res Block& 
            multiscale, MIMO, FAM & 
            L1, FFT\\
            DeepRFT \cite{xint2023freqsel} & 
            MIMO UNet with RFT Block& 
            multiscale, MIMO, FAM & 
            L1, FFT\\
            Uformer \cite{wang2022uformer}& 
            U-shaped ViT& 
            Multi-head WSA, locally-enhanced FFN & 
            L1\\
            Stripformer \cite{tsai2022stripformer}& 
            RED-shaped ViT & 
            Multi-head inter- and intra- strip SA, gated FFN & 
            L1, VGG, Edge\\
            MAXIM-2S  \cite{tu2022maxim} & 
            U-shaped spatially-gated MLP   & 
            Multi-stage, multiscale, MIMO, CSFF, SAM  & 
            L1, FFT\\
            SGAN-B (ours)  & 
            L2 F-Deconv + Res-U GAN-2S & 
            Multi-stage, multiscale, ESA, coord-embd, le-SNR & 
            L1, GAN, MDF\\
            SGAN-E (ours)  & 
            L2 F-Deconv + Res-U GAN-2S &
            Multi-stage, multiscale, ESA, coord-embd, le-SNR & 
            L1, GAN, MDF, VGG\\
            SGAN-F (ours)  & 
            L2 F-Deconv + RFT-U GAN-2S & 
            Multi-stage, multiscale, ESA, coord-embd, le-SNR & 
            L1, GAN, MDF, VGG, FFT\\
            \bottomrule
    	\end{tabular}
        \begin{tablenotes}[para,flushleft]
            \footnotesize
            \vspace{1ex} CA: Channel-wise attention; CGAN: Conditional GAN;  Coord-embd: Coordinate embedding; CSFF: Cross stage feature fusion; ESA: Efficient self-attention; FAM: Feature attention module; FFN: Feed forward net; MLP: Multi-layer perceptron; L2 (F-) Deconv: L2 regularized deconvolution (in feature space);  Le-SNR: Learnable signal-to-noise ratio; MDF: multiscale discriminator feature matching; MIMO: Multi-input and multi-output; REDNet: Residual encoder decoder net; Res-U: Residual UNet; RFT-U: Residual FFT UNet; SAM: Supervised attention module; SA: Self-attention; ViT: Vision Transformer; WSA: Window-based self-attention; 2S: Two-stage.
        \end{tablenotes}
    \end{threeparttable}
\end{table*}

\section{Experiment verification}
\label{sec:Experiment}
To examine whether our SGAN models, trained on simulated images, correctly reconstruct experimentally acquired images, a laboratory prototype \cite{wirth2020half} was used to record the images. The schematic of the setup is shown in Figure \ref{fig:experiment-setup} where a spatial light modulator (SLM) was used to produce the five half-ring phase function. The optical configuration may be described in two parts.
An object (or background scene) illuminated with a red light emitting diode that has been transmitted through a 633 nm laser line filter (10 nm FWHM) is imaged onto a 16-bit CCD camera (SBIG-8300M). A pellicle beam splitter (BS$_1$) is used to superimpose a red HeNe laser beam upon the optical path.  A spatial filter (SF) system is used to remove high spatial frequency artifacts from the laser beam. Lens $L_1$ of focal length $f_1 = 40$ cm collimates both the laser emerging from a 10 $\mathrm{\mu}\text{m}$ pinhole 
in the SF and an arbitrary point on the object. The long focal length ensures that the pupil is nearly uniformly illuminated with the laser beam. The lens $L_2$ of focal length $f_2 = 10$ cm forms an image of the background scene at the field stop (FS) and the diameter of $L_2$ forms the pupil. The pupil is imaged onto the surface of the SLM (Holyeye GAEA) by use of a lens $L_3$ having focal length $f_3 = 10$ cm. The SLM is programmed to holographically imprint the phase function $\phi_R$ onto the reflected optical field.  The reflected light passes through $L_3$ again and then makes a right angle turn owing to the beam spitting cube $BS_2$.  A razor blade $RZ$ is placed in the conjugate plane of the field stop in order to remove undesirable diffracted artifacts caused by the SLM.  Finally, this conjugate plane is relayed to the CCD using the lens $L_4$ of focal length $f_4 = 20$ cm. To record ground truth images without the effects of phase $\phi_R$ and the laser, we removed the laser beam and turned off the SLM. The values of the physical parameters are listed in Table \ref{tab: params_phys}. 
%SLM: Holyeye GAEA; CCD: SBIG-8300M
% to record coded images and ground truth images
% programmed the SLM to produce a constant phase. The values of the physical parameters are listed in Table \ref{tab: params_phys}.

\section{Results and Discussions}
To validate the proposed anti-dazzle imaging scheme, evaluations are conducted in both simulation and experiment. The three variants of our model SGAN-B/E/F are compared in total with 11 alternative image restoration methods qualitatively and quantitatively. Using simulation, the restoration models were also assessed against diverse image contents, illumination conditions, laser strengths and incident angles, as well as sensor noise characterizations. 

\subsection{Baseline Methods and Evaluation Metrics}
The baselines methods include three GANs: Pix2Pix \cite{isola2017image} is the first conditional GAN (CGAN) for domain translation; DeblurGAN \cite{kupyn2018deblurgan} applies CGAN to motion deblurring and replaces L1 loss with VGG loss; ST-CGAN \cite{wang2018stacked} stacks end-to-end two CGANs for shadow detection and removal. Two hybrid methods, which combine L2-regularized deconvolution with DNNs, are also investigated. WienerNet \cite{yanny2022deep} stacks the Wiener deconvolution with a UNet \cite{ronneberger2015u} and learns a universal signal-to-noise ratio. DeepWiener \cite{dong2020deep} applies Wiener deconvolution in feature space followed by a multiscale residual encoder-decoder network for refinement. The remaining six baselines learn the direct mapping from the degraded image to the restored image. Multi-input and multi-output (MIMO)-UNet \cite{cho2021rethinking} is a modified UNet that transforms multiscale input into multiscale output with cross-scale feature attention and fusion.  To make use of the FFT features, DeepRFT \cite{xint2023freqsel} replaces in MIMO-UNet the residual block \cite{he2016deep} with a residual FFT block. MPRNet \cite{zamir2021multi} stacks three UNets with cross-stage feature fusion and attention supervision strategies; Uformer \cite{wang2022uformer} and Stripformer \cite{tsai2022stripformer} are two vision transformers that are built, respectively, on a UNet and a REDNet backbone. They also make use of window-based local attention to reduce the complexity of global self-attention. As an alternative to vision transformers, MAXIM-2S \cite{tu2022maxim} is a U-shaped gated MLP. It employs block and grid partitioning and an axis swapping method to calculate spatial gating \cite{liu2021pay} in linear time. It also extends existing multistage, multiscale, and MIMO strategies with a cross-gating technique. Comparisons of our models with the baselines are summarized in Table \ref{tab:Comp_Baseline-Algo}. 

For the laser-dazzle protection, FFT representations and losses are found to be effective in restoring high-frequency image details. SSIM \cite{wang2004image} and multiscale discriminator feature matching (MDF) losses \cite{wang2018high} appear to have similar but weaker impacts on image structures restoration. L1 loss is shown to encourage in- and outpainting of missing areas, and VGG loss \cite{johnson2016perceptual} improves perceptual qualities. Here we categorize the models as low- and high-performing based on the loss functions. The \textbf{low-performing} models, which trained with combinations of L1, SSIM, and MDF losses, involve MPRNet \cite{zamir2021multi}, Pix2Pix \cite{isola2017image}, ST-CGAN \cite{wang2018stacked}, Uformer \cite{wang2022uformer}, DeepWiener \cite{dong2020deep}, WienerNet \cite{yanny2022deep}, and our SGAN-B. The \textbf{high-performing} models, which make use of FFT and VGG losses, include DeblurGAN \cite{kupyn2018deblurgan}, Stripformer \cite{tsai2022stripformer}, MIMO-UNet \cite{cho2021rethinking}, MAXIM-2S \cite{tu2022maxim}, DeepRFT \cite{xint2023freqsel}, and our SGAN-E/F. Neural networks exhibit a bias towards learning low frequency functions. The use of FFT features in DeepRFT \cite{xint2023freqsel} and our SGAN-F further improves their performances by suppressing the spectral bias. 

The metrics for quantitative evaluations include mean square similarity represented by inverted mean square error (1-MSE); the peak signal-to-noise ratio (PSRN); structural similarity index measure (SSIM) \cite{wang2004image} and multiscale structural similarity index measure (MSSSIM) \cite{wang2003multiscale}; learned perceptual image patch similarity score (1-LPIPS) \cite{zhang2018unreasonable}; and deep image structural and texture similarity (DISTS) \cite{ding2020image}. 

\begin{figure*}[!htb] 
	\begin{center} 
	\includegraphics[width=1.0\linewidth]{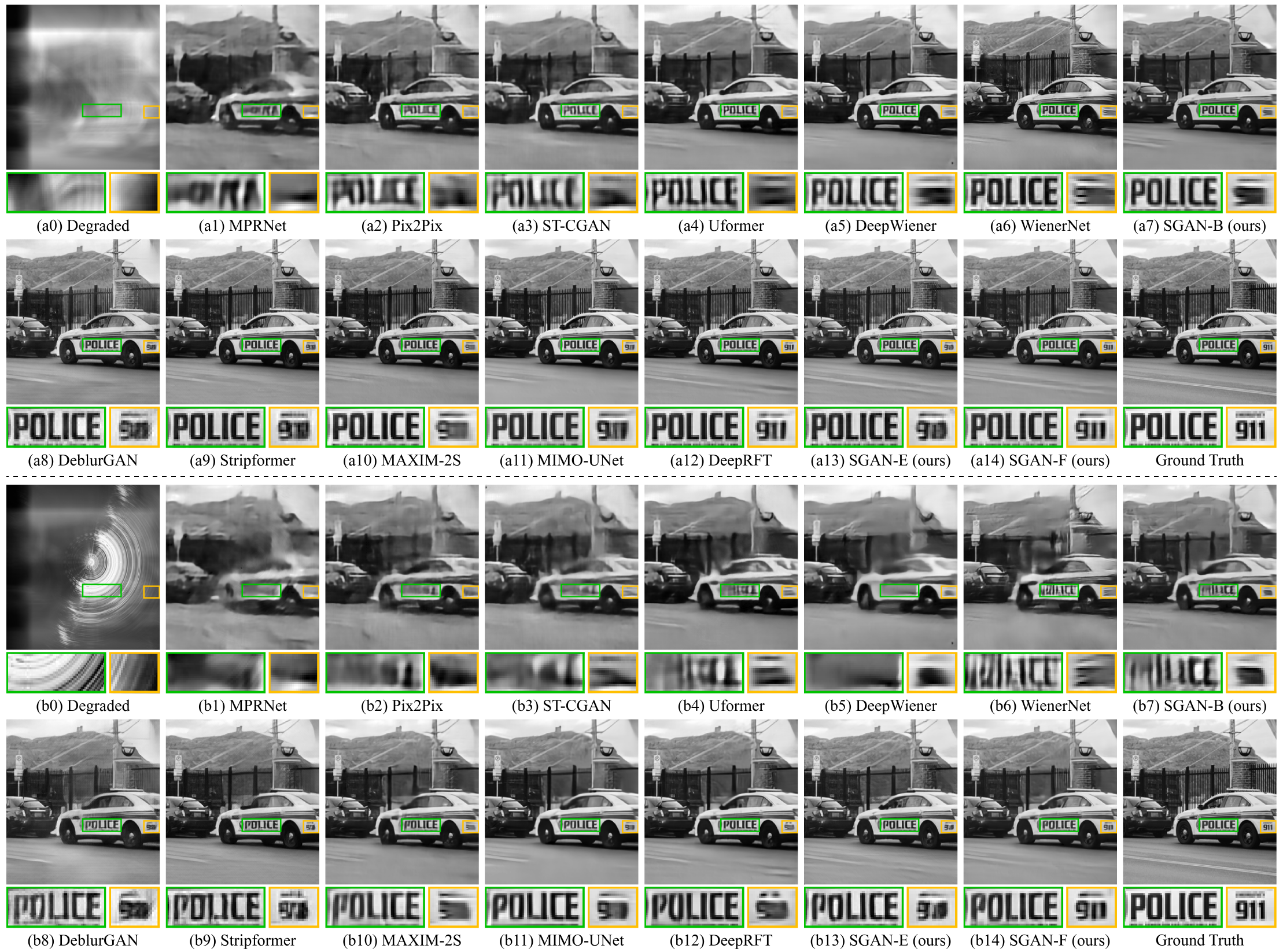} 
	\end{center} 
	\caption{Evaluation of laser-dazzle protection in simulation. Our SGAN-B/E/F models are compared with alternative methods for image restoration of a laser-free case ($\alpha_l = 0$) in rows 1 and 2, and a damaging laser-dazzle case ($\alpha_l = \text{1e6}$) in rows 3 and 4.  In each case, images restored by low- and high-performing models are respectively shown in the top and the bottom strips. MPRNet \cite{zamir2021multi}, Pix2Pix \cite{isola2017image}, and ST-CGAN \cite{wang2018stacked} yield significantly distorted results in both cases. Uformer \cite{wang2022uformer}, DeepWiener \cite{dong2020deep}, WienerNet \cite{yanny2022deep}, and our SGAN-B deliver reasonable recoveries, but perform poorly in the presence of laser dazzle. DeblurGAN \cite{kupyn2018deblurgan}, Stripformer \cite{tsai2022stripformer}, and MAXIM-2S \cite{tu2022maxim} show improvements against laser dazzle in terms of coarse image features; however, fine image details (see the zoom-in image patches outlined by yellow boxes) remain unrecognizable regardless of the laser strengths. Without a laser, high-frequency features become recognizable in the images restored by MIMO-UNet \cite{cho2021rethinking}, DeepRFT \cite{xint2023freqsel}, and our SGAN-E. Among all, our SGAN-F produces the consistently highest fidelity image in both the laser-free and laser-dazzle cases.}\label{fig:resNum}
\end{figure*}

\begin{figure*}[!htb] 
	\begin{center} 
	\includegraphics[width=1.0\linewidth]{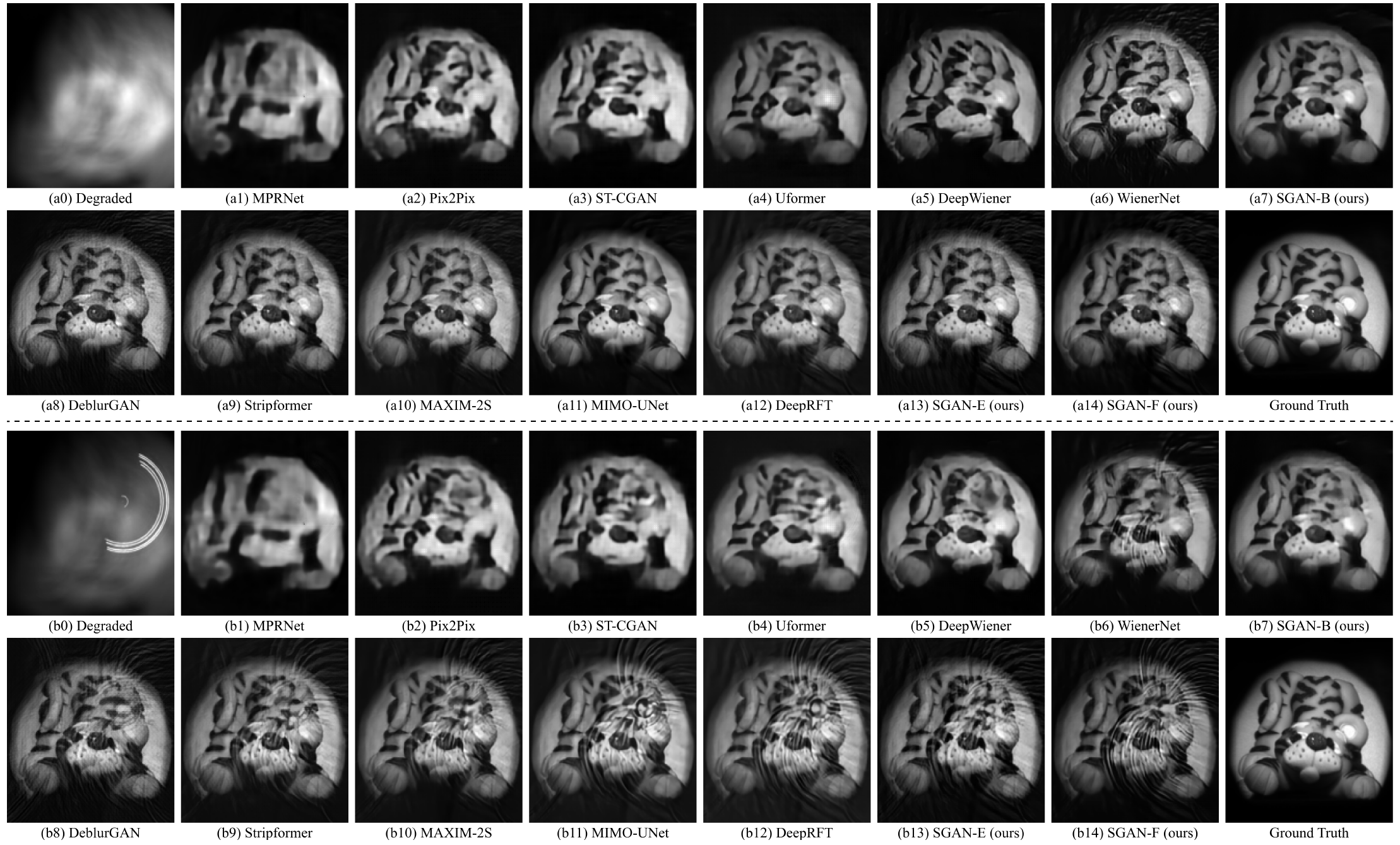} 
	\end{center} 
	\caption{Evaluation of laser-dazzle protection in experiment. Our SGAN-B/E/F models are compared with alternative methods for the restoration of a laser-free scene in rows 1 and 2, and a laser-dazzle scene ($\alpha_l \sim \text{1e4}$) in rows 3 and 4. For each scene, images restored by low- and high-performing models are shown in the top and the bottom strips respectively. Although the high-performing models produce better recoveries of high-frequency image details, the restored images suffer from artifacts and ghost laser diffractive patterns. The performance discrepancies from the numerical cases may be attributed to the laser flare caused by lens internal reflections, a factor not accounted for in the simulation model. The low-performing models produce cleaner results; however, the restored images appear to lack fine image details. In both the laser-free and laser-dazzle cases, our SGAN-B model produces the highest fidelity recovery among all.} \label{fig:resExp}
\end{figure*}

\subsection{Datasets and Training}
\label{ssec: Results_dataset}
A set of 11K unique 8-bit $5120 \times 2880$ color images of versatile contents is collected\cite{unsplash2022}. The images are converted to grayscale as monochromatic scene radiance $b$. The image set is divided into 10K training images and 1K testing images. The coded and uncoded PSFs are generated using Eqs. \ref{eq:PSF_ideal} and \ref{eq:PSF_coded} respectively. During training, coded sensor images are numerically simulated in an online manner, using the physics-based model described in Section \ref{sec:imaging-formation}. To match the resolution of our laboratory camera sensor ($2532 \times 2532$), each image is randomly cropped. The simulated PSF and coded images are then downsampled to $256 \times 256$ using antialiased bicubic interpolation \cite{hu2006adaptive}, to reduce the computational cost of neural network training. Coded images are padded to zero to $384 \times 384$ as input to our SGAN models for the recovery of the boundary areas. Laser strengths $\alpha_l$ are randomly sampled from 10K predetermined values, which are uniformly distributed in the range of $[0,\text{2e6}]$. The incident angles of laser $\vec{n}_l=(n_{u}, n_{v})$ are normally sampled, with the "3-sigma" (three times the standard deviation) set to $0.36\cdot[f/W_s, f/H_s]$ along each axis, where $f$ is the focal length, and $W_s$ and $H_s$ are the sensor width and height respectively. The models are trained with various noise strengths, where the dark current is normally sampled with a standard deviation equal to half its mean $\mu_c = 0.002\text{e}^-$. The read noise is uniformly sampled with $\mu_r \in [350, 400]\text{e}^-$ and $\sigma_r \in [10, 11]\text{e}^-$. The Gaussian-distributed photon noise has coefficients uniformly sampled $c_1 \in [0,25\%]$ and $c_2 \in [0.9, 1.1]$. A variety of background illumination strengths $\alpha_b \in [0.3, 0.7]$ are considered. Exposure times $t$ are normally sampled with a mean of $0.1$ seconds, and the standard deviation is 0.1 times the mean value. 

Adam optimizer with momentum $\beta_1 = 0.9, \beta_2 = 0.999$ is adopted in training. All models are trained for 400 epochs with the learning rates of the weights and biases initialized to 2e-4 and 4e-4 respectively. The learning rates are reduced by half after the first 100 epochs and then by 70 percent every 50 epochs. The biases are initialized to zero. The weights are orthogonally initialized for models other than MIMO-UNet and Stripformer, which are initialized with the Xavier normal. We match the ratio of the L1 loss to the high-performing VGG and FFT losses for models that are trained with them. Other hyperparameters are also tuned to achieve the best performance for each model, and their values are summarized in the Supplementary Information. 

\begin{table*}
	\centering
    \begin{threeparttable}
    	\caption{Quantitative Evaluation of Image Restoration Accuracies (Simulation/Experiment)}
        \label{tab:QuantComp_Results}
    	\setlength{\tabcolsep}{2pt}
    	\begin{tabular}{m{70pt} m{68pt} m{68pt} m{68pt} m{68pt} m{68pt} m{68pt}}
    		\toprule
    		\makecell{Method} &
            \makecell{1-MSE} &
            \makecell{PSNR}&
            \makecell{SSIM}&
            \makecell{MSSSIM} &
            \makecell{1-LPIPS}&
            \makecell{DISTS}\\
    		\midrule
            MPRNet \cite{zamir2021multi} & 
            \makecell{0.9872/0.9894} &  
            \makecell{19.45/20.75} &  
            \makecell{0.5938/0.5623} &  
            \makecell{0.7759/0.8112} &  
            \makecell{0.5356/0.6785} &  
            \makecell{0.6595/0.7182}\\ 
            Pix2Pix \cite{isola2017image} &  
            \makecell{0.9914/0.9945} &  
            \makecell{21.34/23.59} &  
            \makecell{0.6829/0.6582} &  
            \makecell{0.8602/0.9157} &  
            \makecell{0.6129/0.7517} &  
            \makecell{0.7122/0.7671}\\ 
            ST-CGAN \cite{wang2018stacked} &   
            \makecell{0.9920/0.9947} &   
            \makecell{21.68/23.79} &   
            \makecell{0.6921/0.7220} &   
            \makecell{0.8732/0.9187} &   
            \makecell{0.6219/0.7555} &   
            \makecell{0.7164/0.7685}\\ 
            WienerNet \cite{yanny2022deep} &   
            \makecell{0.9920/0.9945} &   
            \makecell{21.80/23.66} &   
            \makecell{0.7539/0.6433} &   
            \makecell{0.8819/0.9220} &   
            \makecell{0.6902/0.7057} &   
            \makecell{0.7679/0.7536}\\ 
            DeepWiener \cite{dong2020deep} &   
            \makecell{0.9936/0.9948} &   
            \makecell{22.73/23.99} &   
            \makecell{0.7393/\underline{0.7270}} &   
            \makecell{0.8994/\underline{0.9372}} &   
            \makecell{0.6630/0.8022} &   
            \makecell{0.7507/0.8009}\\ 
            Uformer \cite{wang2022uformer} &   
            \makecell{0.9932/0.9947} &   
            \makecell{22.57/23.75} &   
            \makecell{0.7308/0.6687} &   
            \makecell{0.9070/0.9348} &   
            \makecell{0.6675/\underline{0.8047}} &   
            \makecell{0.7482/\underline{0.8063}}\\ 
            SGAN-B (ours)  &   
            \makecell{0.9959/\textbf{0.9968}} &   
            \makecell{24.77/\textbf{26.03}} &   
            \makecell{0.7981/\textbf{0.7950}} &   
            \makecell{0.9397/\textbf{0.9622}} &   
            \makecell{0.7206/\textbf{0.8526}} &   
            \makecell{0.7810/\textbf{0.8418}}\\ 
            DeblurGAN \cite{kupyn2018deblurgan}&    
            \makecell{0.9942/0.9947} &  
            \makecell{23.25/23.80} &   
            \makecell{0.7958/0.6196} &   
            \makecell{0.9188/0.9199} &   
            \makecell{0.7688/0.6471} &   
            \makecell{0.8163/0.7003}\\ 
            MIMO-UNet \cite{cho2021rethinking} &   
            \makecell{0.9958/0.9952} &   
            \makecell{24.81/\underline{24.85}} &   
            \makecell{0.8443/0.6508} &   
            \makecell{0.9502/0.9338}&   
            \makecell{0.7870/0.7538} &   
            \makecell{0.8282/0.7801}\\ 
            Stripformer \cite{tsai2022stripformer} &   
            \makecell{0.9965/0.9955} &   
            \makecell{25.68/24.66} &   
            \makecell{0.8351/0.6407} &   
            \makecell{0.9508/0.9365} &   
            \makecell{0.8133/0.7165} &   
            \makecell{\underline{0.8539}/0.7357}\\ 
            MAXIM-2S \cite{tu2022maxim} &   
            \makecell{0.9968/0.9949} &   
            \makecell{26.16/24.06} &   
            \makecell{0.8419/0.6357} &   
            \makecell{0.9519/0.9306} &   
            \makecell{0.7802/0.7343} &   
            \makecell{0.8225/0.7487}\\ 
            DeepRFT \cite{xint2023freqsel} &   
            \makecell{0.9972/0.9930} &   
            \makecell{\underline{26.76}/22.77} &   
            \makecell{0.8609/0.5843}&   
            \makecell{0.9635/0.9123} &  
            \makecell{ 0.8034/0.6890} &   
            \makecell{0.8354/0.7287}\\ 
            SGAN-E (ours)  &   
            \makecell{\underline{0.9972}/\underline{0.9957}} &   
            \makecell{26.71/24.80} &   
            \makecell{\underline{0.8647}/0.6697} &   
            \makecell{\underline{0.9648}/0.9352} &   
            \makecell{\textbf{0.8332}/0.7084} &   
            \makecell{\textbf{0.8689}/0.7479}\\ 
            SGAN-F (ours)  &   
            \makecell{\textbf{0.9977}/0.9915} &   
            \makecell{\textbf{27.67}/22.13} &   
            \makecell{\textbf{0.8691}/0.6410} &   
            \makecell{\textbf{0.9676}/0.9044}&   
            \makecell{\underline{0.8173}/0.6908} &   
            \makecell{0.8483/0.7346}\\ 
            \bottomrule
    	\end{tabular}
        \begin{tablenotes}[para,flushleft]
            \vspace{0.5ex} The \textbf{highest} and the \underline{second highest} scores are highlighted. For the numerical evaluations, our SGAN-F outperforms all other models in terms of MSE, PSNR, SSIM, and MSSIM, while our SGAN-E achieves the highest perceptual (LPIPS) and texture (DISTS) qualities. Our SGAN-B not only produces the best restoration of the experimental data but also outperforms all other low-performing models (Uformer, DeepWiener, WienerNet, ST-CGAN, Pix2Pix, and MPRNet) in the restoration of the numerically simulated images. Compared to numerical results, high-performing models (DeblurGAN, MIMO-UNet, Stripformer, MAXIM-2S, DeepRFT, and our SGAN-E/F) underperform in the reconstruction of experimental images, which is potentially explained by the simulation model not accounting for lens flare. 
        \end{tablenotes}
    \end{threeparttable}
\end{table*}

\subsection{Evaluation in Simulation}
Qualitative evaluation of the anti-dazzle imaging is shown in Figure \ref{fig:resNum}, where the proposed SGAN-B/E/F models are compared with alternative methods for image restoration. Background illumination strength $\alpha_b = 0.7$, coefficients of photon noise $c_1 = 20\%$ and $c_2 = 1.0$, dark current noise $\mu_c = 0.002 \text{e}^-$, as well as the read noise $\mu_r = 390 \text{e}^-$ and $\sigma_r = 10.5 \text{e}^-$ remain the same in simulation. Rows 1 and 2 showcase the restoration of a laser-free scene ($\alpha_l = 0$). Restorations of a scene from potentially damaging laser dazzle ($\alpha_l = 1\text{e}6$) are shown in rows 3 and 4. For each scence, the degraded image and the images restored by low-performing models are presented at the top strip; the ground truth image and the images restored by high-performing models are shown at the bottom strip. To demonstrate comparisons in detail, the coarse- and fine-scale image features of each image are highlighted by green and yellow colored boxes respectively. MPRNet \cite{zamir2021multi}, Pix2pix \cite{isola2017image}, and ST-CGAN \cite{wang2018stacked} tend to perform poorly regardless of the laser strengths. The images produced by these methods appear to be highly distorted, which makes them hardly reconizable. Uformer \cite{wang2022uformer}, DeepWiener \cite{dong2020deep}, WienerNet \cite{yanny2022deep}, and our SGAN-B generate reasonable reconstructions of the laser-free scene. However, a significant amount of image features remain distorted and unrecognizable in the laser-dazzle case. DeblurGAN \cite{kupyn2018deblurgan}, Stripformer \cite{tsai2022stripformer}, and MAXIM-2S \cite{tu2022maxim} produce recognizable recovery on the coarse scale, but fine image details are barely captured. Although DeepRFT \cite{xint2023freqsel}, MIMO-UNet \cite{cho2021rethinking}, and our SGAN-E further improve the restoration accuracies, the fine image details remain unrecognizable in the laser-dazzle case. In both the laser-free and laser-dazzle cases, our SGAN-F outperforms all other image restoration models and produces the consistently highest-fidelity reconstructions for the anti-dazzle imaging. 

Quantitative evaluations are performed on a set of 7K test images simulated from a thousand ground truth scenes and seven laser strengths $\alpha_l = \{0, 10^k | k = 1, 2, ..., 6\}$. Other parameters follow the sampling scheme of the training set (see Section \ref{ssec: Results_dataset}). Restoration accuracies are averaged across the entire set for each of the six metrics and compared in Table \ref{tab:QuantComp_Results}. Quantitative and qualitative evaluations align, with high-performing models producing more accurate restorations than low-performing ones. Our SGAN-F demonstrates superiority over all other models in terms of MSE, PSNR, SSIM, and MSSIM, while our SGAN-E achieves the best perceptual (LPIPS) and texture (DISTS) qualities. Our SGAN-B not only outperforms other low-performing models in all metrics but also exceeds the high-performing DeblurGAN in MSE, PSNR, SSIM, and MSSIM. Both qualitative and quantitative results demonstrate the effectiveness of FFT representations and losses in restoring high-frequency information. The results also reaffirm the role of VGG loss in improving the perceptual qualities of the restored images. 

\begin{figure*}[ht] 
	\begin{center} 
	\includegraphics[width=1.0\linewidth]{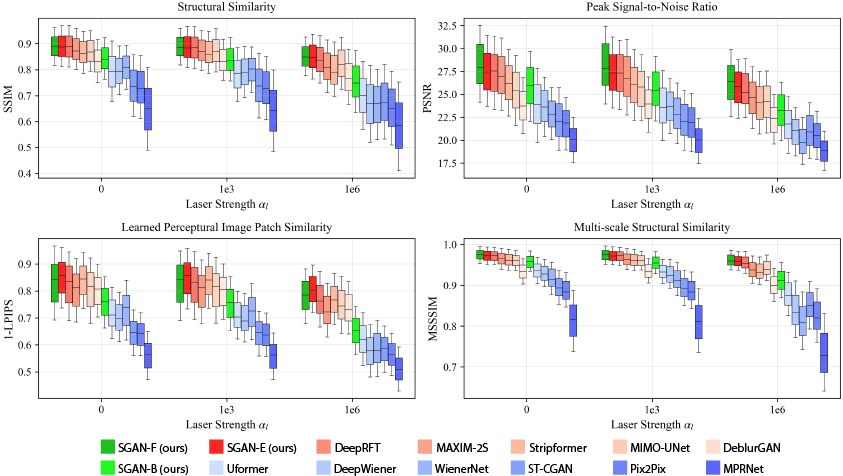} 
	\end{center} 
	\caption{Robust analysis of laser dazzle protection against laser strengths $\alpha_l = 0, 1\text{e}3$, and $1\text{e}6$. The top/middle/bottom bar of each box represents the range in which 25/50/75 percentile of the accuracies falls, with the top and bottom whiskers indicating the minimum and maximum respectively. Our SGAN-F/E show stronger tolerance to laser dazzle compared to other models across all four metrics, with SGAN-E leading SGAN-F in LPIPS. Our SGAN-B exceeds other low-performing models by a large margin in all cases and outperforms the high-performing DeblurGAN \cite{kupyn2018deblurgan} in PSNR and MSSSIM.}\label{fig:resStats}
\end{figure*}

\subsection{Experimental Results}
We demonstrate the generalization of our anti-dazzle imaging to the experiment. A set of two images of the same background scene was acquired using the prototype described in Section \ref{sec:Experiment}. One image is captured for the laser-free scene ($\alpha_l = 0$) and the other is obtained for the laser-dazzled scene. The laser strength in the latter case is adjusted to ten times the system saturation threshold ($\alpha_l \sim 1\text{e}4$).  

Qualitative comparisons of the proposed SGAN models with alternative methods are provided in Figure \ref{fig:resExp}. The restored images of a laser-free scene are showcased in rows 1 and 2. For the scene where sensor-damaging laser dazzle is presented, the restored images are shown in rows 3 and 4. In each case, images restored by low- and high-performing models are shown respectively in top and bottom strips. Compared to low-performing models, high-performing models produce better recovery of high-frequency image information. Contrary to the numerical results, however, the image restored by the high-performing models suffers from artifacts and ghost laser diffractive patterns in the laser-dazzle case. Such discrepancies may be attributed to the laser flare caused by the inter-reflections in the lens, a factor not accounted for in the simulation pipeline. The low-performing WienerNet also exhibits similar sensitivity due to the use of SSIM loss. For both the laser-free and laser-dazzle scenes, low-performing models produce cleaner recoveries at the expense of losing fine image details. The image restored by MPRNet is barely recognizable. In the absence of a laser, Uformer \cite{wang2022uformer}, DeepWiener \cite{dong2020deep}, and WienerNet \cite{yanny2022deep} yield reasonable restorations despite observable distortions. Our SGAN-B produces the highest quality recovery among the low-performing models, which is consistent with the numerical results. 

The quantitative evaluation of the experimental results is shown in the Table \ref{tab:QuantComp_Results}, where the mean restoration accuracy of each model is summarized. Our SGAN-B achieves the highest restoration accuracies for all six metrics, which aligns with the qualitative comparisons shown in Figure \ref{fig:resExp}.  Models that follow closely include DeepWiener \cite{dong2020deep} in SSIM and MSSIM, Uformer \cite{wang2022uformer} in LPIPS and DISTS, MIMO-UNet \cite{cho2021rethinking} in PSNR, and our SGAN-E in MSE evaluation. Similarly to what was observed in qualitative evaluations, the underperformance of the expectedly high-performing models is also evidenced in quantitative measures. 

\subsection{Robustness Analysis}
\label{ssec: Stats_Analysis}
To evaluate the robustness of our SGAN models against laser dazzle, quantitative evaluations are performed on a set of 3K test images, which are simulated from a thousand ground truth scenes and three laser strengths $\alpha_l = 0, 1\text{e}3$, and $1\text{e}6$. The SSIM, PSNR, LPIPS, and MSSIM scores are averaged respectively for each laser strength.  The box plots of the four metrics are shown in Figure \ref{fig:resStats}. Each box indicates the interquartile range of the restoration accuracies, with the top/middle/bottom bar representing the 25/50/75 percentiles. The top and bottom whiskers of each box indicate the min and max accuracies respectively. Our SGAN-F outperforms DeepRFT \cite{xint2023freqsel} by a larger margin in PSNR regardless of laser strength. Compared to DeepRFT, our SGAN-F also exhibits a higher tolerance to laser dazzle, evidenced by less significant reductions in SSIM and MSSIM as the laser strength increases. Our SGAN-E scores the highest in LPIPS, reaffirming the effectiveness of VGG loss in improving perceptual quality. Additionally, our SGAN-B outperforms other low-performing models by a large margin across all four metrics, with its advantage becoming increasingly evident as the laser strength climbs up. Our SGAN-B also outperforms the high-performing DeblurGAN \cite{kupyn2018deblurgan} in PSNR and MSSIM. Evaluations of robustness against varying degrees of degradation stages are demonstrated in Supplementary Information, including increased strengths of laser and noise, and reduced background illuminations. 

\subsection{Discussions}
In summary, we introduced a novel anti-dazzle system using computational imaging approach, whereby PSF engineered phase mask is combined with deep learning based image restoration algorithm. Both our simulated and experimental results suggest that the proposed anti-dazzle imaging scheme has the potential to protect the camera sensor from being damaged by laser radiations, without compromising the image quality when the scene is laser-free, i.e., under normal operating conditions.  When a laser is directed onto the detector, the image is nearly fully recovered using our neural sandwich GAN (SGAN) models. 
The foregoing numerical results showcase the restoration of high-fidelity images of complex scenes using our models, particularly the SGAN-F where the
laser dazzle has a peak irradiance as high as $10^6$ times the sensor saturation threshold. Compared to state-of-the-art image restoration methods, the proposed SGAN-F model also demonstrated improved restoration quality and robustness to a wide range of imaging conditions, such as background illuminations, noise, and time-varying laser strengths and positions. 

Although our experiment compared well in some cases, especially when the dazzle strength was weak, ghost images attributed to the lens internal reflection were found to introduce noise in the reconstructed images when the dazzle strength was high. Fine tuning the numerically trained model on a set of experimental images provides a way to mitigate the discrepancies.  In cases where ground truth images are impractical to acquire (e.g., drone uses, weather conditions), unsupervised domain adaptation methods \cite{liang2023comprehensive} may be employed to transfer the model learning from numerical to experimental data. 

Future work is required to extend this approach to broadband illumination and a variety of laser wavelengths. Examination of other unfavorable imaging conditions, such as shift-variant distortions, motion blur, atmosphere, and adverse weathers, are also indispensable towards practical applications. Taking advantage of spectral representations in neural networks may be worth further investigation to develop fully blind image restoration without compromising the quality of the restored images. Aliasing issue presented in image downsampling may be further mitigated using learnable subpixel sampling techniques \cite{kim2022mssnet}. Provided limited computational resources, deep learning based super-resolution may allow restoration of high-resolution images from degraded inputs of low resolutions. Furthermore, a fully differentiable end-to-end imaging pipeline \cite{tseng2021differentiable} that jointly optimizes the phase function and image restoration algorithm may further improve the performance of the antidazzle imaging system.

\section{Conclusion}
Our neural Sandwich GAN (SGAN) technique, combined with a wavefront-coded phase mask producing a five half-ring point spread function, has been demonstrated to protect an imaging sensor from laser dazzle or damage. Unlike nonlinear optical approaches, our optical mechanism is governed by the response of linear lossless materials and is, in principle, broadband and instantaneous. The PSF associated with the phase-coded aperture function is generated almost instantaneously, providing immediate sensor protection. The response time of the entire system is therefore determined by post-processing, where the video rate (25 FPS) is achieved for the restoration of $256 \times 256$ images using our SGAN models. In experiment and simulation, we achieved respectively an irradiance dynamic range of $10^4$ and $10^6$ times the sensor saturation threshold. What is more, the system was trained to perform across a wide range of laser strengths and incident angles, background illumination conditions, and sensor noise characteristics. Multiple uses of our anti-dazzle system are envisioned, including protection the sensors of autonomous vehicles, consumer and security cameras, HDR imaging, and laser safety head-mounted displays. The proposed SGAN frameworks may also find applications in other computational imaging systems that suffer from arbitrary sensor saturation as well as boundary cutoffs and the image blur caused by an extended PSF of large support (e.g., lensless cameras). 

\section{Acknowledgement}
We thank Kyle Novak (NRL), and Jacob Wirth, Prateek Srivastava, and Arnab Ghosh (RIT) for valuable discussions on computational imaging and experimental verification. This work is supported by the Funding. U.S. ONR N00014-19-1-2520 and N0001422WX00932. 
\bibliographystyle{IEEEtran}
\bibliography{ref_intro, ref_related_coded, ref_related_dmethod, ref_related_gmethod, ref_result}

\end{document}

\title{Supplementary Information: \\ Learning to See Through Dazzle}
\author{Xiaopeng Peng, Erin F. Fleet, Abbie T. Watnik, Grover A. Swartzlander, Jr.}

\IEEEdisplaynontitleabstractindextext
\IEEEpeerreviewmaketitle
\maketitle

\noindent In this document we provide additional discussion and results in support of the primary text.

\section{Noise Calibration}
Here we discuss details of sensor calibration and noise estimates. In the experiment, an SBIG-8300M camera was used. The parameters obtained from the manufacture specification \cite{sbig8300} include the full well capacity $e_{\text{sat}}$, sensor gain $\mathcal{G}$, pixel pitch $\Delta x$, the mean of the dark current $\mu_c$, and the bit depth per channel bpc. The values of these parameters are listed in Table 1 in the primary text. 

\begin{figure*}[!htb] 
	\begin{center} 
	\includegraphics[width=1.0\linewidth]{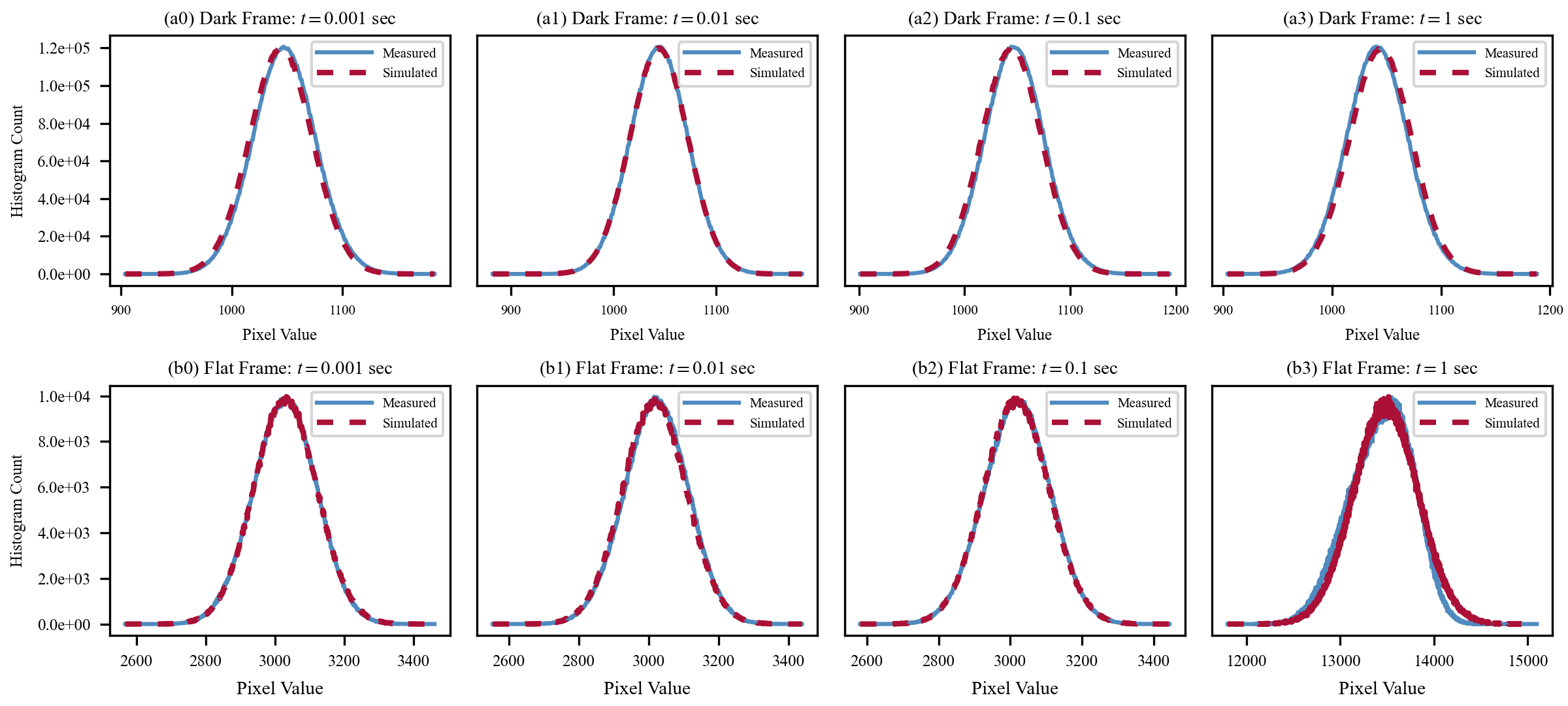} 
	\end{center} 
	\caption{Histogram of measured and simulated dark frames (row a) and gray frames (row b). Columns 0-3: The histograms calculated for exposure times $t = 0.001, 0.01, 0.1$ and 1 second respectively. The simulation agrees with the experiment in most cases. A slight discrepancy is observed in the case of a gray frame at an exposure time of one second.}\label{fig: noise_hist}
\end{figure*}

The read noise in the simulation is considered to be Gaussian distributed $n_r \sim \mathcal{N}(\mu_r, \sigma_r)$, and the values of its mean $\mu_r$ and standard deviation $\sigma_r$ vary with exposure time. Here, we calibrate these values from a set of dark frames that were taken at a set of exposure times $t = \{10^k, k =-3, ...0\}$. For each exposure time, 20 frames are captured with the lens cap closed and averaged to a single dark frame. The value of $\mu_r$ is given by the mean of the bias frame, which is the dark frames taken at the minimum exposure time $t = 0.001$. The value of $\sigma_r$ is given by the standard deviation of the dark frame at each particular exposure time. To validate the estimates, we compare the histogram of the measured dark frames with that of the simulated dark frames in Figure \ref{fig: noise_hist} (a0) - (a3). The simulations appear to agree with the measurements. 

To validate the noise estimates in the presence of photons, a gray patch is displayed on an LED monitor. By imaging that patch without attaching a lens, a set of gray frames is captured at the same set of exposure times $t = \{10^k, k =-3, ...0\}$. The distance between the camera and the screen is carefully adjusted to best avoid lens falloff. Numerical gray frames are simulated using our physics-based image formation model (see Section 3 in the primary text). Gaussian-distributed photon noise with modulation coefficients $c1 = 0.2$ and $c_2 = 1.0$ is found to best represent our sensor in practice. Similarly, the histograms of the simulated and measured gray frame pairs are compared in Figures \ref{fig: noise_hist} (b0) - (b3). The histogram of simulated gray frames agrees with the histogram of the measured gray frames for shorter exposure times, but a small discrepancy is observed as the exposure time increases to one second.

\section{Net Architecture}

The architecture of our SGAN-B/E/F models is inspired by multiple methods, including UNet \cite{ronneberger2015u}, residual net \cite{he2016deep}, Pix2PixHD \cite{wang2018high}, DeepWiener \cite{dong2020deep}, efficient self-attention \cite{shen2021efficient}, and DeepRFT \cite{xint2023freqsel}. Let {\fontfamily{pcr}\selectfont C} denote the \textbf{Convolution Block}, which has a Conv2D-InstanceNorm-ReLU layout with a stride of 1.  The \textbf{Residual Block} {\fontfamily{pcr}\selectfont R} consists of two consecutive Conv2D-InstanceNorm-ReLU layers with a stride of 1. A skip connection is established to add the input to the output. The \textbf{Encoder Block} and \textbf{Decoder Block} are denoted as {\fontfamily{pcr}\selectfont E} and {\fontfamily{pcr}\selectfont D} respectively, which have the Conv2D-InstanceNorm-ReLU structure with a stride of 2. The \textbf{Self-Attention block} 
 {\fontfamily{pcr}\selectfont A} consists of four Conv2D layers, each has a stride size of 1 and a kernel size of 1.  The architectures described below are denoted as a combination of the block name and the number of kernels. The basic residual blocks used in our SGAN-B and SGAN-E models are replaced with the \textbf{Residual FFT block} (see Figure 2 (a)) in the SGAN-F model to make use of the FFT representation. Similarly, the convolutional layers in the discriminators $D_1$ and $D_2$ are replaced by \textbf{FFT Convolutional block} (see Figure 2(b)) SGAN-F. 

\begin{figure}[!htb] 
	\begin{center} 
	\includegraphics[width=0.8\linewidth]{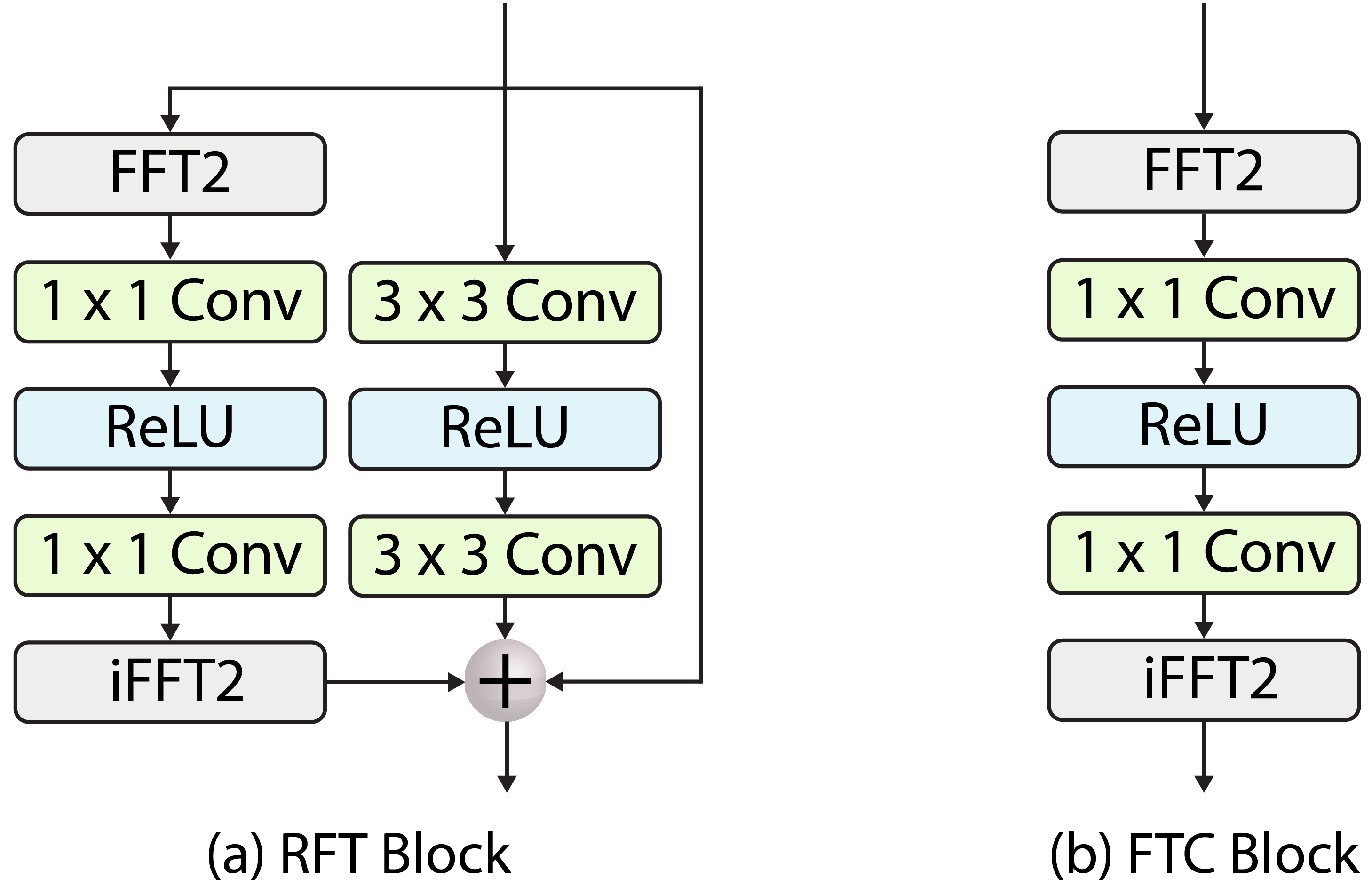} 
	\end{center} 
	\caption{Illustration of the Fourier representations in our SGAN-F model. (a) Residual FFT (RFT) block and (b) FFT Convolutional (FTC) block}
	\label{fig: blocks}
\end{figure}

\subsection{Pre and Post Restoration Generator}
\label{appendix:CGAN}
The generators $G_1$ and $G_2$ of the SGAN share the same residual UNet architecture, in which an encoder sequence is followed by a decoder sequence.\\
\noindent \textbf{Encoder sequence}:\\
{\small\fontfamily{pcr}\selectfont C-ERRA64-ERRA128-ERRA256-ERRA512-}\\
{\small\fontfamily{pcr}\selectfont ERRA512-ERRA512-ERRA512} \\
\noindent \textbf{Decoder sequence}:\\
{\small\fontfamily{pcr}\selectfont DRRA512-DRRA512-DRRA512-} \\
{\small\fontfamily{pcr}\selectfont DRRA512-DRRA256-DRRA128-DRRA64}\\
\noindent where the convolution block {\small\fontfamily{pcr}\selectfont C} has a kernel size of 7, the encoder block {\small\fontfamily{pcr}\selectfont E} and the decoder block {\small\fontfamily{pcr}\selectfont D} have a kernel size of 4, and the residual blocks {\small\fontfamily{pcr}\selectfont R} have a kernel size of 3. Skip connections concanate activations from the layer $i$ to the layer $n-i$ in each generator.  

\subsection{Feature Extractor}
\label{appendix:FE}
\noindent The feature extractor $FE$ has a structure of: {\small\fontfamily{pcr}\selectfont CRRR16}, where the convolutional block {\small\fontfamily{pcr}\selectfont C} and the residual blocks in this module have a kernel size of 5.

\subsection{Discriminators}
\label{appendix:D}
The discriminators $D_1$ and $D_2$ share the same Markovian (patch) discriminator, the layout of which is given by: {\small\fontfamily{pcr}\selectfont E64-E128-E256-E512}. Each encoder block {\small\fontfamily{pcr}\selectfont E} has a kernel size of 4. The output patch has a dimension of $70 \times 70$.

\section{Hyper-parameters in Training}
In addition to the use of FFT features, training models with different loss functions is also found to significantly affect image restoration accuracies in anti-dazzle imaging. FFT losses tend to encourage the restoration of high-frequency image details and thus produce finer and sharper restored images. SSIM loss \cite{wang2004image} and multiscale discriminator feature matching (MDF) loss \cite{wang2018high} exhibit similar but weaker impacts on the restoration of image structures. VGG loss \cite{johnson2016perceptual} improves the perceptual quality of the recovered image, and L1 loss appears to encourage the restoration of saturated areas. We match the ratio of L1, FFT, and VGG losses for the models trained with them. Other hyperparameters are also fine-tuned to achieve the best performance for each model. The values of the loss hyperparameters are listed in Table \ref{tab:Hyperparameters}.

\begin{table*}
	\centering
    \begin{threeparttable}
    	\caption{Values of Hyperparameters in Model Training}
        \label{tab:Hyperparameters}
    	\setlength{\tabcolsep}{2pt}
        \begin{tabular}{m{70pt} m{70pt} m{40pt} m{40pt} m{40pt} m{40pt} m{40pt} m{40pt} m{40pt}}
            \toprule
            \makecell{Model} &
            \makecell{$\lambda_{L1}$}&
            \makecell{$\lambda_{EDGE}$}&
            \makecell{$\lambda_{FFT}$}&
            \makecell{$\lambda_{VGG}$}&
            \makecell{$\lambda_{SSIM}$}&
            \makecell{$\lambda_{MDF}$}&
            \makecell{$\lambda_{ADV}$} &
            \makecell{$\lambda_{GP}$}\\
            \midrule
            MPRNet \cite{zamir2021multi}           & 1.0 & 0.05 & - & - & -   & - & -   & -  \\ 
            Pix2Pix \cite{isola2017image}          & 100 & -    & - & - & -   & - & 1.0 & 10 \\
            ST-CGAN \cite{wang2018stacked}         & s1:100, s2:100 & -    & - & - & -   & - & 1.0 & 10 \\ 
            WienerNet \cite{yanny2022deep}         & 1.0 & -    & - & - & 1.0 & - & -   & - \\
            DeepWiener \cite{dong2020deep}         & 1.0 & -    & - & - & -   & - & -   & -  \\  
            Uformer \cite{wang2022uformer}         & 1.0 & -    & - & - & -   & - & -   & - \\ 
            DeblurGAN \cite{kupyn2018deblurgan}    & 1.0 & - & - & - & - &- &1.0&0.01\\ 
            MIMO-UNet \cite{cho2021rethinking}     & 1.0 & - & 10 & - & - &- &-&-\\ 
            Stripformer \cite{tsai2022stripformer} & 1.0 & - & 10 & - & - &- &-&-\\ 
            MAXIM-2S \cite{tu2022maxim}            & 1.0 & - & 10 & - & - &- &-&-\\
            DeepRFT \cite{xint2023freqsel}         & 1.0 & - & 10 & - & - &- &-&-\\
            SGAN-B (ours)                           & s1:200, s2:50 & - & - & 50 & - &- &1.0&10\\
            SGAN-E (ours)                           & s1:300, s2:50 & - & - & 100 & - &100 &0.1&1.0\\
            SGAN-F (ours)                           & s1:10, s2:10 & - & 100 & 0.1 & - &0.1 &0.1&1.0\\
            \bottomrule
        \end{tabular}
        \begin{tablenotes}[para,flushleft]
            \footnotesize
            \vspace{0.5ex} s1: stage 1; s2: stage 2
        \end{tablenotes}
    \end{threeparttable}
\end{table*}
 
\section{Additional Evaluations}
In addition to the assessment of anti-dazzle imaging in simulation as presented in the primary text (see Section 6.3) where our proposed SGAN models are compared with the 11 alternative baseline algorithms for image restoration, here we extend our evaluation to an additional background scene. Furthermore, the robustness of anti-dazzle imaging against diverse and challenging imaging conditions is demonstrated using two distinct scenes. In the robust analysis, we compare the performance of our SGAN-F models with two high-performing algorithms and showcase its superior tolerance to adverse imaging conditions.

\begin{figure*}[!htb] 
	\begin{center} 
	\includegraphics[width=1.0\linewidth]{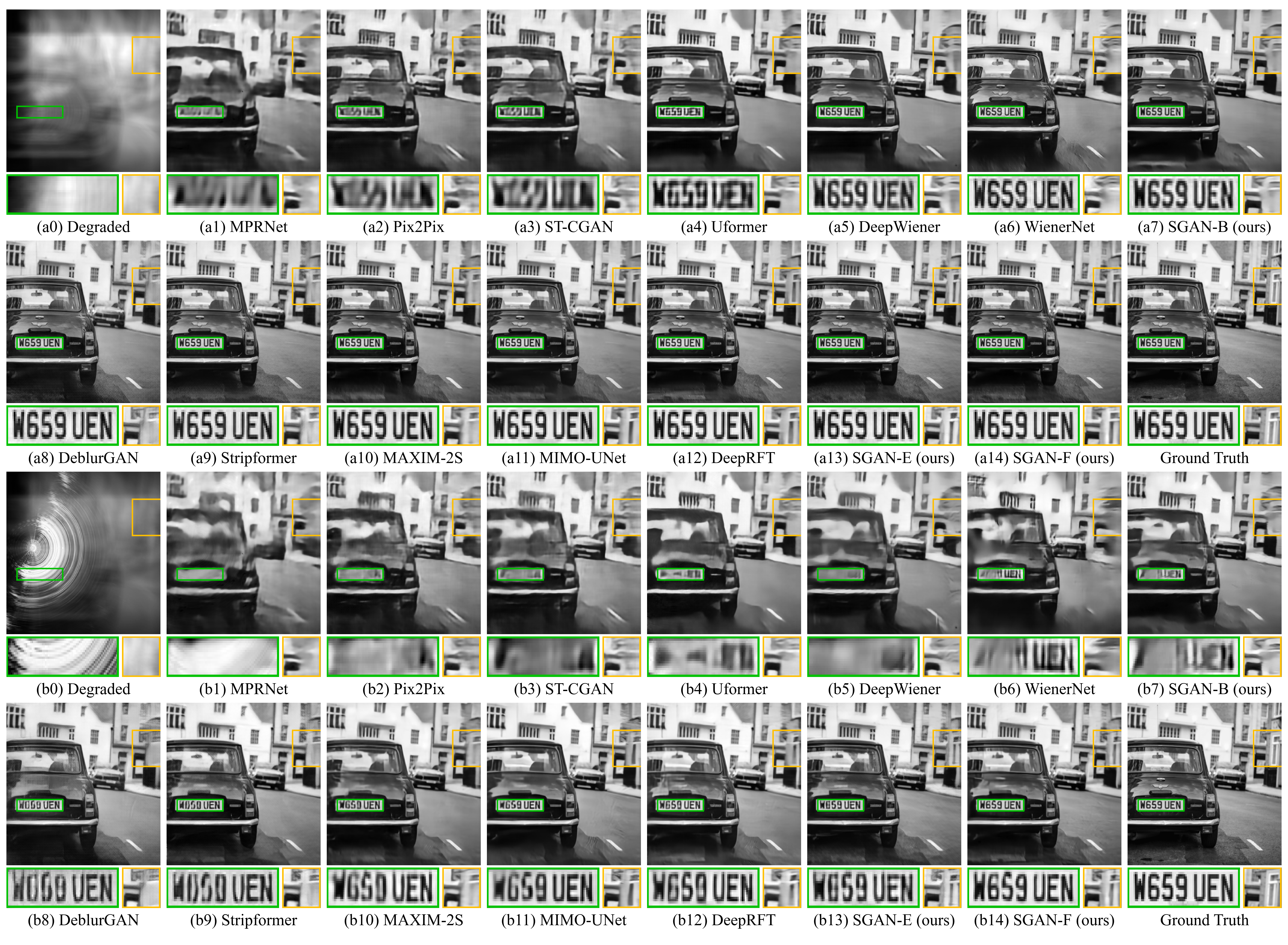} 
	\end{center} 
	\caption{Evaluation of laser-dazzle protection in simulation. Our SGAN-B/E/F models are compared with alternative methods for image restoration of a laser-free case ($\alpha_l = 0$) in rows 1 and 2, and a damaging laser-dazzle case ($\alpha_l = \text{1e6}$) in rows 3 and 4.  In each case, images restored by low- and high-performing models are respectively shown in the top and the bottom strips. MPRNet \cite{zamir2021multi}, Pix2Pix \cite{isola2017image}, and ST-CGAN \cite{wang2018stacked} yield significantly distorted results in both cases. Uformer \cite{wang2022uformer}, DeepWiener \cite{dong2020deep}, WienerNet \cite{yanny2022deep}, and our SGAN-B deliver reasonable recoveries, but perform poorly in the presence of laser dazzle. DeblurGAN \cite{kupyn2018deblurgan}, Stripformer \cite{tsai2022stripformer}, and MAXIM-2S \cite{tu2022maxim} show improvements against laser dazzle in terms of coarse image features; however, fine image details (see the zoom-in image patches outlined by yellow boxes) remain unrecognizable regardless of the laser strengths. Without a laser, high-frequency features become recognizable in the images restored by MIMO-UNet \cite{cho2021rethinking}, DeepRFT \cite{xint2023freqsel}, and our SGAN-E. Among all, our SGAN-F produces the consistently highest fidelity image in both the laser-free and laser-dazzle cases.}
	\label{fig: result1}
\end{figure*}

Qualitative evaluation of the anti-dazzle imaging is shown in Figure \ref{fig: result1}, where the proposed SGAN-B/E/F models are compared with alternative methods for image restoration. Background illumination strength $\alpha_b = 0.7$, coefficients of photon noise $c_1 = 20\%$ and $c_2 = 1.0$, dark current noise $\mu_c = 0.002 \text{e}^-$, as well as the read noise $\mu_r = 390 \text{e}^-$ and $\sigma_r = 10.5 \text{e}^-$ remain the same in simulation. Rows 1 and 2 showcase the restoration of a laser-free scene ($\alpha_l = 0$). Restorations of a scene from potentially damaging laser dazzle ($\alpha_l = 1\text{e}6$) are shown in rows 3 and 4. For each scence, the degraded image and the images restored by low-performing models are presented at the top strip; the ground truth image and the images restored by high-performing models are shown at the bottom strip. To demonstrate comparisons in detail, the coarse- and fine-scale image features of each image are highlighted by green and yellow colored boxes respectively. MPRNet \cite{zamir2021multi}, Pix2pix \cite{isola2017image}, and ST-CGAN \cite{wang2018stacked} tend to perform poorly regardless of the laser strengths. The images produced by these methods appear to be highly distorted, which makes them hardly reconizable. Uformer \cite{wang2022uformer}, DeepWiener \cite{dong2020deep}, WienerNet \cite{yanny2022deep}, and our SGAN-B generate reasonable reconstructions of the laser-free scene. However, a significant amount of image features remain distorted and unrecognizable in the laser-dazzle case. DeblurGAN \cite{kupyn2018deblurgan}, Stripformer \cite{tsai2022stripformer}, and MAXIM-2S \cite{tu2022maxim} produce recognizable recovery on the coarse scale, but fine image details are barely captured. Although DeepRFT \cite{xint2023freqsel}, MIMO-UNet \cite{cho2021rethinking}, and our SGAN-E further improve the restoration accuracies, the fine image details remain unrecognizable in the laser-dazzle case. In both the laser-free and laser-dazzle cases, our SGAN-F outperforms all other image restoration models and produces the consistently highest-fidelity reconstructions. 

\begin{figure*}[!htb] 
	\begin{center} 
	\includegraphics[width=1.0\linewidth]{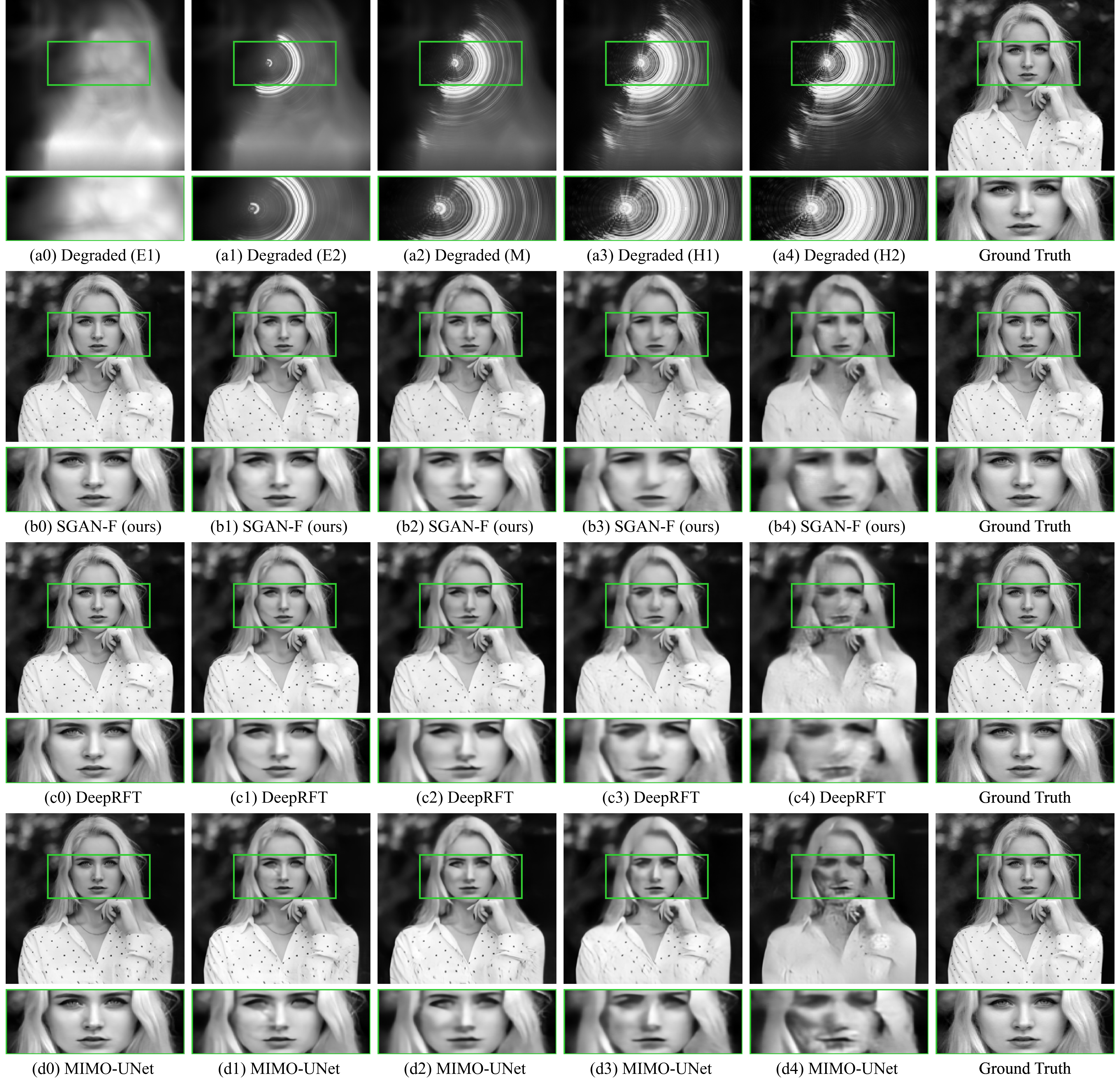} 
	\end{center} 
	\caption{Evaluation of the the robustness of laser dazzle protection against five degradation stages: easiest (E1), easy (E2), medium (M), hard (H1) and hardest (H2) from columns 1-5. Compared to the ground truth image shown in column 6, the imaging conditions deteriorate as laser dazzle and photon noise increase, and the strength of background illumination decrease. A set of phase-coded images (row 1) illustrates the progressive worsening of image quality through these stages. The image restoration performance of our SGAN-F (row 2) is compared with that of DeepRFT (row 3) and MIMO-UNet (row 4). Although all models produce high-quality recoveries of the laser-free scene, DeepRFT and MIMO-UNet showcase a significant increase in distortion from E1 to H2. In contrast, our SGAN-F model consistently delivers higher fidelity restorations at every stage, demonstrating its superior robustness against the challenging imaging condition.}
	\label{fig:result2}

\end{figure*}

\begin{figure*}[!htb] 
	\begin{center} 
	\includegraphics[width=1.0\linewidth]{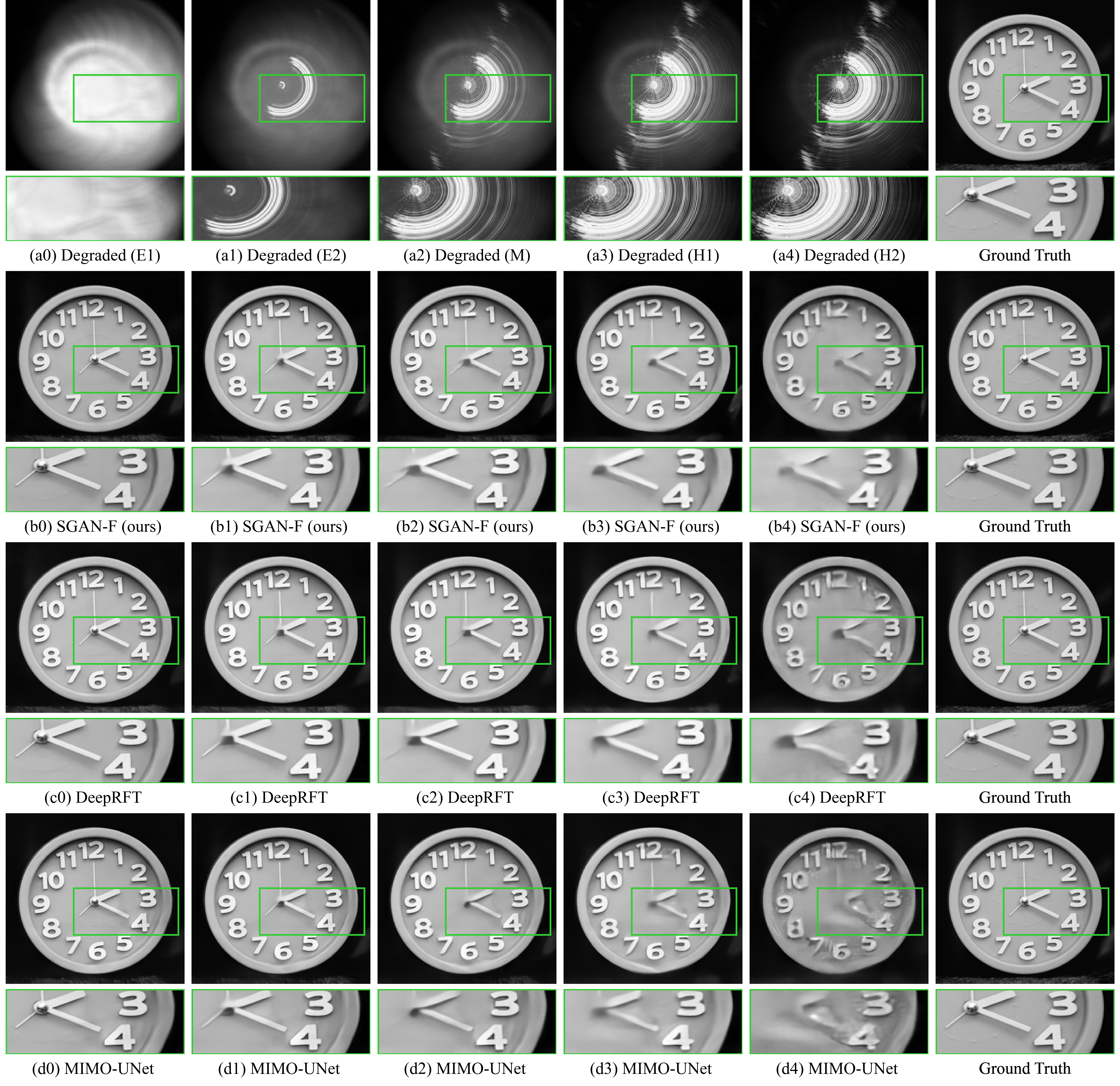} 
	\end{center} 
	\caption{Evaluation of the the robustness of laser-dazzle protection against five degradation stages: easiest (E1), easy (E2), medium (M), hard (H1) and hardest (H2) from columns 1-5. Compared to the ground truth image shown in column 6, the imaging conditions deteriorate as laser dazzle and photon noise increase, and the strength of background illumination decrease. A set of phase-coded images (row 1) illustrates the progressive worsening of image quality through these stages. The image restoration performance of our SGAN-F (row 2) is compared with that of DeepRFT (row 3) and MIMO-UNet (row 4). Although all models produce high-quality recoveries of the laser-free scene, DeepRFT and MIMO-UNet showcase a significant increase in distortion from E1 to H2. In contrast, our SGAN-F model consistently delivers higher fidelity restorations at every stage, demonstrating its superior robustness against the challenging imaging condition.}
	\label{fig:result3}
\end{figure*}

The robustness of the anti-dazzle imaging is evaluated across five levels of degradation: easiest (E1), easy (E2), medium (M), hard (H1), and hardest (H2). Following the progression of degradation stages from the least to the most challenging conditions, the laser strength increases from zero to a potentially damaging level $\alpha_l$ = 1.5\text{e} 6, the coefficient of photon noise raises from $c_1 = 1\%$ to $20\%$, and the background illumination strength decreases from $\alpha_b = 0.8$ to 0.2. To demonstrate the generalization of image restoration models under low light conditions, the background illumination strengths used in the testing are set to exceed its training limits by $10\%$ at the lower and upper ends. The values of these degradation parameters at each stage are listed in Table \ref{tab:Degradation_modes}. The performance of our SGAN-F model is compared in Figure \ref{fig:result2} with two high-performing baseline methods: DeepRFT \cite{xint2023freqsel} and MIMO-UNet \cite{cho2021rethinking}). Degraded images at different stages are shown in rows (a0)-(a4). Correspondingly, the images restored by SGAN-F are presented in (b0)-(b4), restored images by DeepRFT in (c0)-(c4), and those by MINO-UNet in (d0)-(d4). The same ground truth image is shown in the last column of each row. All three models produce high-fidelity restoration at E1, the least challenging condition. At E2 and M, artifacts and distortions are observed in images restored by DeepRFT and MIMO-UNet, while the images restored by our SGAN-F barely changed. At H1, the DeepRFT and MIMO-UNet results become significantly distorted. Although the images restored by our SGAN-F at this stage appear to be less sharp, a relatively higher degree of fidelity is maintained. DeepRFT and MIMO-UNet perform poorly in H2, where the restored scenes are highly distorted and hardly recognizable. Our SGAN-F remarkably preserves the structural integrity of the scene at this stage. The robustness evaluation of a second scene is shown in Figure \ref{fig:result3}. As expected, the additional results are consistent with those for the first scene. At all degradation stages, our SGAN-F model consistently outperforms DeepRFT and MIMO-UNet in capturing fine detail of the scene and restoring high-fidelity images. 

\begin{table}[tb]
\begin{threeparttable}
\centering
\captionsetup{justification=centering}
\caption{Values of Parameter in Robustness Evaluation}
\label{tab:Degradation_modes}
\begin{tabular}{m{30pt} m{30pt} m{30pt} m{30pt} m{30pt} m{30pt}}
\toprule
\makecell{Param} &
\makecell{E1} &
\makecell{E2}&
\makecell{M}&
\makecell{ H1} &
\makecell{H2}\\
\midrule
\makecell{$\alpha_l$} & 
\makecell{0} & 
\makecell{3e4} & 
\makecell{3e5} & 
\makecell{1e6} & 
\makecell{1.5e6}\\
\makecell{$\alpha_b$} & 
\makecell{0.8} & 
\makecell{0.6} & 
\makecell{0.4} & 
\makecell{0.3} & 
\makecell{0.2}\\ 
\makecell{$c_1$} &  
\makecell{$1\%$} &  
\makecell{$3\%$} &  
\makecell{$5\%$} &  
\makecell{$10\%$} &  
\makecell{$20\%$}\\
\bottomrule
\end{tabular}
\begin{tablenotes}[para,flushleft]
%\footnotesize
\vspace{0.5ex} Param: parameter; E1: easiest; E2: easy; M: medium; H1: hard; H2: hardest. Values of other parameters are fixed in the testings, they include $c_2=1.0$, $\mu_r = 390\mathrm{e}^-$, $\mu_r = 10.5\mathrm{e}^-$, and $\mu_c = 0.002\mathrm{e}^-$.
\end{tablenotes}
\end{threeparttable}
\end{table}

\bibliographystyle{IEEEtran}
\bibliography{Supplement}